\documentclass{article}

\usepackage[main,final]{neurips_2026}
\usepackage[utf8]{inputenc} % allow utf-8 input
\usepackage[T1]{fontenc}    % use 8-bit T1 fonts
\usepackage{hyperref}       % hyperlinks
\usepackage{url}            % simple URL typesetting
\usepackage{booktabs}       % professional-quality tables
\usepackage{amsfonts}       % blackboard math symbols
\usepackage{nicefrac}       % compact symbols for 1/2, etc.
\usepackage{microtype}      % microtypography
\usepackage{xcolor}         % colors
\usepackage{amsmath}
\usepackage{booktabs}
\usepackage{algorithm}
\usepackage{algpseudocode}
\usepackage{wrapfig}
\usepackage{subcaption}
\usepackage[dvipsnames]{xcolor}
\usepackage{adjustbox}
\usepackage{tabularx} % Include in your preamble for tabularx support
\usepackage{times}
\usepackage{latexsym}
\usepackage{amssymb} 
\usepackage[T1]{fontenc}
\usepackage[most]{tcolorbox}
\tcbuselibrary{skins, breakable}
\usepackage[utf8]{inputenc}
\usepackage{float}
\usepackage{multirow}
\usepackage{microtype}

\usepackage{inconsolata}
\usepackage{subcaption} 
\usepackage{graphicx}
\usepackage{enumitem}
\usepackage{soul}
\usepackage{cleveref}
\usepackage{graphicx}

\usepackage[table]{xcolor}

\title{When Form Changes but Logic Doesn’t: Building Logic-invariant LLMs through Logical Structure}

\author{%
    Xuyuan Liu$^{1}$ \hspace{2em} Xinshuai Dong$^{2}$ \hspace{2em} Elynn Chen$^{3}$ \hspace{2em} Yujun Yan$^{1}$\\
    $^{1}$Dartmouth College \hspace{2em}
    $^{2}$Carnegie Mellon University \hspace{2em}
    $^{3}$New York University\\
    \texttt{\{xuyuan.liu.gr, yujun.yan\}@dartmouth.edu} \\
    \texttt{xinshuad@andrew.cmu.edu} \qquad
    \texttt{elynn.chen@stern.nyu.edu}
}

\begin{document}

\maketitle

\begin{abstract}

Large language models (LLMs) have achieved strong performance across diverse reasoning tasks, yet it remains unclear whether this performance reflects an understanding of underlying logical structure or a reliance on shallow semantic cues. A key indicator of such understanding is whether models treat logically equivalent inputs consistently, producing the same predictions despite differences in expression. Through systematic evaluation on curated data, we show that even strong LLMs often fail this test: they do not consistently preserve predictions across logically equivalent inputs and generalize poorly to unseen logical forms.
To address this limitation, we propose \textbf{LoGI}cal \textbf{ST}ructure-guided reasoning (LoGIST), a framework that guides LLM reasoning with explicit logical structure. 
LoGIST maps each input instance to a decision diagram, a directed acyclic graph that captures the underlying logical structure, and encodes this graph into a logical embedding that conditions the LLM's reasoning process.
Across multiple training settings, LoGIST reduces inconsistency under logical equivalence and improves generalization to unseen logical forms, suggesting that logical invariance is a useful target for building more robust and generalizable logical reasoning in LLMs.
\footnote{Our code is available at \href{https://github.com/GraphmindDartmouth/LoGIST}{\textcolor{ForestGreen}{https://github.com/GraphmindDartmouth/LoGIST}}.}
\end{abstract}

\section{Introduction}

Large language models (LLMs) have demonstrated remarkable performance across a wide range of reasoning tasks, from mathematical problems to logical puzzles~\cite{DBLP:conf/nips/Wei0SBIXCLZ22,DBLP:conf/iclr/ZhouSHWS0SCBLC23,DBLP:conf/iclr/0002WSLCNCZ23,zeng2026sagemitigatinglonghorizonreasoning, yan2020neural}. However, whether LLMs' reasoning capabilities reflect a genuine understanding of underlying logic or reliance on semantic cues remains unclear. Several recent studies suggest that, in some settings, LLMs may rely on surface-level patterns~\citep{DBLP:conf/iclr/MirzadehASTBF25,DBLP:conf/naacl/ZhouZAKZ24,DBLP:conf/emnlp/YuanZZZL24} and memorized world knowledge~\citep{DBLP:conf/emnlp/YangSWQZWRWL25,DBLP:conf/acl/DuSSWSC24,DBLP:conf/iclr/BerglundTKBSKE24,DBLP:conf/iclr/0003AEAAZW25,DBLP:conf/acl/LiuCDLZWYCC26} rather than formal logical structures. A central test of logical understanding is whether models capture logical invariance: logically equivalent inputs should yield the same conclusions despite differences in expression. 
{In this paper, we show that even state-of-the-art LLMs often fail this test. Using our curated dataset of logically equivalent input variants, we find that advanced LLMs exhibit substantial performance drops and frequent decision flips when the equivalent underlying logic is expressed in different forms. These failures reveal weak logical consistency and suggest that current LLMs remain overly sensitive to surface form rather than shared logical structure. We further show that addressing this invariance gap improves generalization to unseen logical forms.}

{Prior work on improving LLMs’ logical reasoning has largely emphasized task-level accuracy over learning reusable logical structure. Some methods couple LLMs with external symbolic tools, such as logic solvers or theorem provers~\citep{pan-etal-2023-logic,DBLP:conf/nips/YeCDD23,olausson2023linc,DBLP:journals/tmlr/PoesiaGZG24}; while effective, they rely on external systems for formal reasoning, leaving the LLM’s intrinsic structure-aware reasoning less examined. Other methods rely on data-driven training~\citep{qi2025large,DBLP:conf/nips/SaparovPPJKK023,DBLP:conf/icml/MorishitaMYS23,DBLP:conf/nips/ZhouP0CCLCZMZ24}, but do not explicitly model logical structure. As a result, models may learn distribution-specific patterns rather than reusable logical principles, limiting generalization to unseen logical forms.}

{To address this gap, we propose \textbf{LoGI}cal \textbf{ST}ructure-guided reasoning (LoGIST). LoGIST maps each logical reasoning question to a decision diagram, a directed acyclic graph that captures the input’s underlying logical structure independently of its surface form. Logically equivalent input variants are mapped to the same decision diagram. LoGIST then encodes these decision diagrams into structure-aware logical embeddings through contrastive learning and uses the embeddings to condition the LLM’s reasoning. Across diverse training strategies, LoGIST improves both logical consistency across equivalent inputs and generalization to unseen logical forms, suggesting that logical invariance is a useful target for building more robust and generalizable logical reasoning in LLMs. Our further analysis shows that LoGIST learns to encode logical composition as a linear direction, thus explaining its generalizability.}

In summary, our contributions are as follows:

\begin{itemize}[leftmargin=1.0em, itemsep=1.5pt, topsep=2pt, parsep=0pt, partopsep=0pt]
    \item[-]\textbf{New Insight}: {We identify a key limitation of current LLMs: even strong models often fail to reason consistently over logically equivalent inputs, revealing weak logical invariance and limited generalization to unseen logical forms.}

     \item[-] \textbf{New Method}: 
     {We propose LoGIST, a framework that maps natural-language inputs to decision diagrams and encodes their logical structure into embeddings that condition LLM reasoning, enabling more consistent predictions across surface-form variants.}
    
    \item[-] \textbf{New Perspective}:
    {We show that logical invariance and generalization are closely connected: explicitly modeling logical structure helps LLMs reason more consistently over equivalent inputs and generalize better to unseen logical forms.}
  
\end{itemize}

\vspace{-6pt}
\section{Preliminaries}
\vspace{-6pt}

{We begin by introducing the notation and key concepts for formal reasoning. We then discuss rules of replacement, which produce logically equivalent reformulations while preserving truth conditions. Finally, we present decision diagrams as representations that encode logical structure beyond surface-form differences.}

We study formal reasoning tasks in which each instance consists of a premise set and a query. We denote each instance by $d=(\Gamma,\mathcal{Q})$, where $\Gamma=\mathcal{F}\cup\mathcal{R}$ is a finite premise set consisting of a fact set $\mathcal{F}$ and a rule set $\mathcal{R}$, and $\mathcal{Q}$ is the query whose truth value is to be determined under $\Gamma$.
% Each instance has \yy{a paired} representations, a formal logical form and a natural-language realization, therefore written as
{Each instance contains a paired formal and natural-language representation:}
\(
d=\big((\Gamma^{\mathrm{FL}},\mathcal{Q}^{\mathrm{FL}}),\;(\Gamma^{\mathrm{NL}},\mathcal{Q}^{\mathrm{NL}})\big),
\)
where $(\Gamma^{\mathrm{FL}},\mathcal{Q}^{\mathrm{FL}})$ {is the formal logical form and} $(\Gamma^{\mathrm{NL}},\mathcal{Q}^{\mathrm{NL}})$ {is its natural-language realization, obtained through a verbalization function} $\mathbb{T}$, such that
\(
(\Gamma^{\mathrm{NL}},\mathcal{Q}^{\mathrm{NL}}) = \mathbb{T}(\Gamma^{\mathrm{FL}},\mathcal{Q}^{\mathrm{FL}}).
\)
%For dataset construction\yy{don't say for dataset construction... sounds like you are introducing your data construction process. this section should be preliminaries, nothing about what you do, what customization you do, it's about what the domain knowledge is. so directly say that the formal premise set can be further decomposed into...}, we further partition 
% \xy{The formal premise set can be further decomposed as $\Gamma^{\mathrm{FL}} = \mathcal{F} \cup \mathcal{R}$, where $\mathcal{F}$ denotes the set of facts and $\mathcal{R}$ denotes the set of rules.} Each fact $f_i \in \mathcal{F}$ is a ground literal, that is, an atomic formula or its negation, specifying a property of an entity or a relation between entities. For example, the fact \texttt{Carlos is good at soccer} can be expressed as $\text{GoodAtSoccer}(\texttt{Carlos})$  \yy{give one example here} Each rule $r_j \in \mathcal{R}$ is a quantified first-order sentence constructed from the logical connectives $(\land, \lor, \neg, \to)$ and quantifiers $(\forall, \exists)$, expressing the conditions under which certain properties or relations hold.For example 
% \yy{give an example here.}
{Each fact $f_i \in \mathcal{F}$ is a ground literal, \textit{i.e.}, an atomic formula or its negation containing no variables, specifying a property of an entity or a relation among entities. For example, the fact \textit{Carlos is good at soccer} is expressed as $\text{GoodAtSoccer}(\texttt{Carlos})$, where $\text{GoodAtSoccer}(\cdot)$ is a predicate and $\texttt{Carlos}$ is a constant denoting a specific entity. Each rule $r_j \in \mathcal{R}$ is a first-order logical statement 
that specifies how one or more conditions imply a conclusion. Rules are built from predicates, logical connectives such as $\land, \lor, \neg, \to$, and, when applicable, quantifiers such as $\forall, \exists$.
% built from logical connectives ($\land, \lor, \neg, \to$) and, when applicable, quantifiers ($\forall, \exists$), expressing relationships between facts and their conclusions. 
{For example, the rule \textit{if someone is good at soccer, then they join the soccer club} can be written as $\forall x (\text{ GoodAtSoccer}({x}) \to \text{JoinSoccerClub}({x}))$, where $x$ is a variable ranging over individuals.} }
% Given a consistent premise set $\Gamma^{\mathrm{FL}}$, 
%the task is to determine the logical status of the query $\mathcal{Q}^{\mathrm{FL}}$. Formally, the label is \textbf{True} if $\Gamma^{\mathrm{FL}} \models \mathcal{Q}^{\mathrm{FL}}$, \textbf{False} if $\Gamma^{\mathrm{FL}} \models \neg \mathcal{Q}^{\mathrm{FL}}$, and \textbf{Uncertain} if neither entailment holds.
% the task is to predict the logical status of $\mathcal{Q}^{\mathrm{FL}}$ under $\Gamma^{\mathrm{FL}}$ as one of three labels: \textbf{True} if $\Gamma^{\mathrm{FL}} \models \mathcal{Q}^{\mathrm{FL}}$, \textbf{False} if $\Gamma^{\mathrm{FL}} \models \neg \mathcal{Q}^{\mathrm{FL}}$, and \textbf{Uncertain} otherwise, where $\models$ denotes classical logical entailment. 
{Given a consistent premise set $\Gamma^{\mathrm{FL}}$, the task is to determine whether a query $\mathcal{Q}^{\mathrm{FL}}$ logically follows from the premises. The label is \textbf{True} if the premises necessarily imply the query ($\Gamma^{\mathrm{FL}} \models \mathcal{Q}^{\mathrm{FL}}$), \textbf{False} if they necessarily imply its negation ($\Gamma^{\mathrm{FL}} \models \neg \mathcal{Q}^{\mathrm{FL}}$), and \textbf{Uncertain} if neither conclusion follows. Here, $\models$ denotes classical logical entailment, meaning truth preservation from the premises to the conclusion.}

\begin{wrapfigure}{r}{0.35\linewidth}
\centering
\vspace{-0.6cm}
\includegraphics[width=\linewidth]{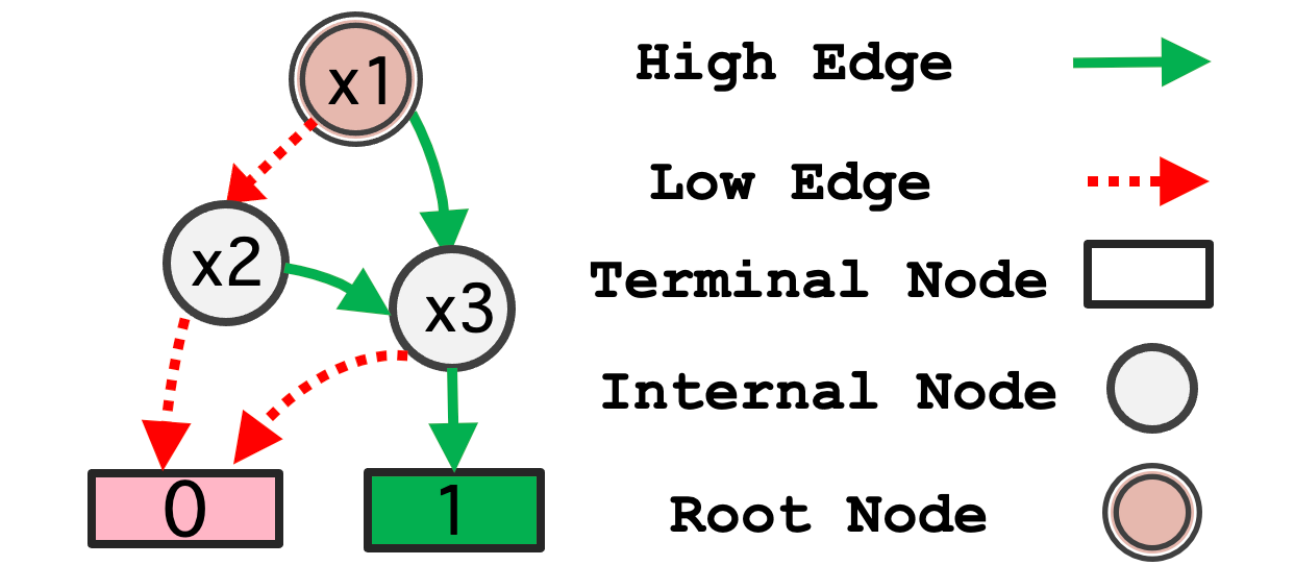}
\vspace{-0.3cm}
\caption{\small ROBDD example for function $F:(x_1 \vee x_2) \wedge x_3$ with variable order $x_1 \prec x_2 \prec x_3 $.}
    \label{fig:robdd_example}
    \vspace{-0.8cm}
\end{wrapfigure}

\paragraph{Rule of Replacement} The Rule of Replacement~\citep{10.5555/579402,10.5555/2481002,DBLP:books/daglib/0082098,DBLP:journals/logcom/Fitting02} allows any subformula to be replaced by a logically equivalent one while preserving the truth value of the full formula. Examples include De Morgan's laws, commutativity, material implication, and other standard equivalence laws~\citep{copi1990introduction}, which alter syntactic form while preserving logical meaning.

 \paragraph{Reduced Ordered Binary Decision Diagram (ROBDD)} An ROBDD~\cite{DBLP:journals/tc/Bryant86,DBLP:journals/csur/Bryant92} is a rooted directed acyclic graph representing a Boolean function. Let $F:\{0,1\}^m \to \{0,1\}$ be a Boolean function defined over an ordered variable list $(x_1,\ldots,x_m)$, 
 {where each variable $x_i$ corresponds to the truth value of a ground atom.}
 % where each variable $x_i$ corresponds to a ground atom. 
 An ROBDD contains a root node $\rho$, two terminal nodes $\tau_0$ and $\tau_1$ labeled with outputs $0$ and $1$, and internal nodes associated with Boolean variables. Each internal node labeled $x_i$ has a low branch for $x_i=0$ and a high branch for $x_i=1$. We denote the sets of low- and high-branch edges by $E_0$ and $E_1$, respectively. Given a Boolean function $F$ and a fixed variable order, an ordered BDD recursively branches on variables in that order until the function value is determined. The ROBDD is obtained by merging isomorphic subgraphs and removing nodes whose low and high successors are identical. Under a fixed variable ordering, logically equivalent formulas over the same variables induce the same ROBDD, yielding a canonical graph representation. We present an example ROBDD for the function $F:(x_1 \vee x_2) \wedge x_3$ with variable order $x_1 \prec x_2 \prec x_3 $ in Fig.~\ref{fig:robdd_example}. 

\vspace{-5pt}

\section{Challenge: LLMs Struggle with Logic-invariant Reasoning}
\vspace{-5pt}

\begin{figure}
    \centering
    \includegraphics[width=0.98\linewidth]{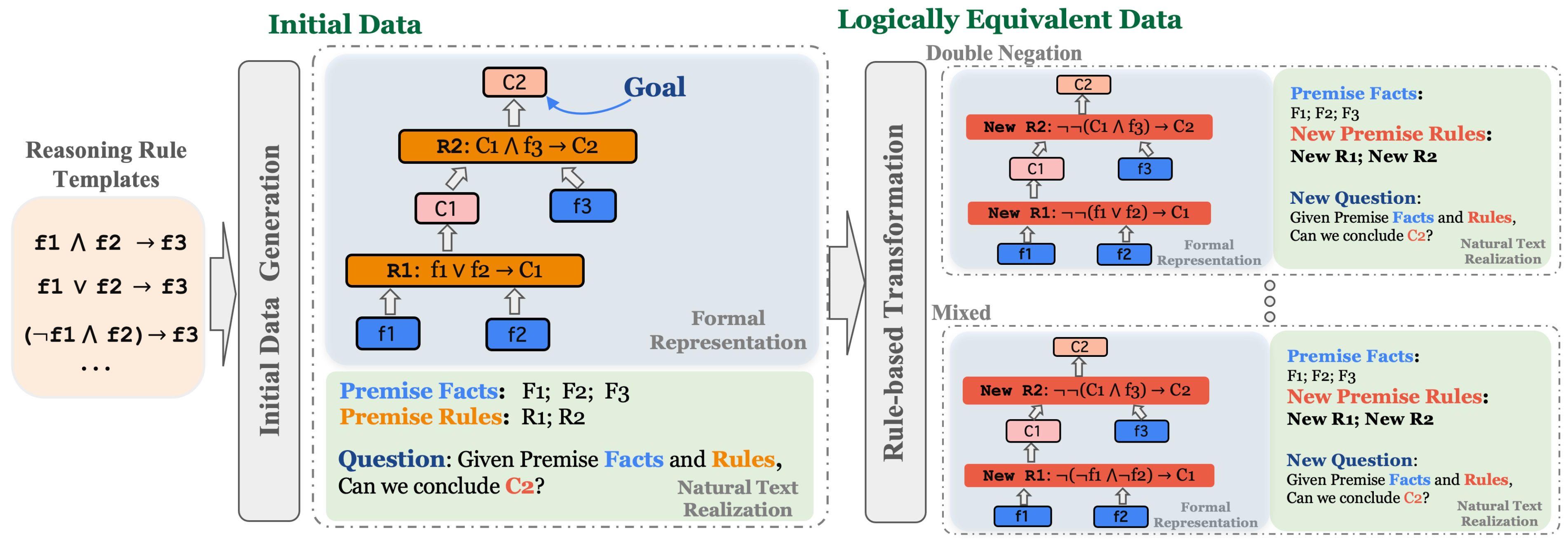}
    \caption{\small{{
    Pipeline for generating logically equivalent data. 
    We instantiate reasoning problems from rule templates and apply rules of replacement to the premise rules while keeping the fact set and query fixed. This produces surface-form variants that are logically equivalent and share the same ground-truth label.
    }}}
   
    \vspace{-6pt}
    \label{fig:data_transfomation}
\end{figure}

It remains unclear whether current LLMs base their predictions on the logical structure of the input, \textit{i.e.}, the rules and dependencies that determine the query’s truth value, or on surface-level cues such as entities, relations, and wording. To systematically examine this question, we construct a dataset by transforming each instance into multiple logically equivalent variants with different surface expressions. The full data generation pipeline is illustrated in Fig.~\ref{fig:data_transfomation}. We then evaluate whether LLMs make invariant predictions across these variants.

\subsection{Constructing Logically Equivalent Data}

\label{subsec:data_curation}

\paragraph{Data Initialization}
We start by constructing a reasoning dataset in which each example contains both a formal representation and its corresponding textual realization. Building on ProverGen~\citep{qi2025large}, we define a step-by-step deductive process where each step applies a grounded implicational rule to derive a new fact. The derived fact can either serve as a premise for subsequent reasoning steps or constitute the final answer. 
Using this procedure, we build the base dataset \(\mathbb{D}_{\text{base}}\). Each example \(d_i \in \mathbb{D}_{\text{base}}\) contains a pair of formal premises and query, \((\Gamma^{\mathrm{FL}}_i,\mathcal{Q}^{\mathrm{FL}}_i)\), together with its corresponding natural-language realization, \((\Gamma^{\mathrm{NL}}_i,\mathcal{Q}^{\mathrm{NL}}_i)\), which serves as the input to LLMs for reasoning.

\paragraph{Rule-Based Transformation under Logical Equivalence}
\label{subsec:rule_based}
From \(\mathbb{D}_{\text{base}}\), we construct logically equivalent variants by applying \textit{rules of replacement} to the formal rule sets while keeping the fact sets fixed. Specifically, for each example \(d_i \in \mathbb{D}_{\text{base}}\), we apply a transformation \(\phi\) to its rule set \(\mathcal{R}_i\), yielding
$\Gamma^{\mathrm{FL}}_{i,\text{new}} = (\mathcal{F}_i,\,\phi(\mathcal{R}_i))$.
This transformation preserves the original facts and deductive dependency while changing the expression of the rules. We consider three replacement rules: \textit{material implication},  \textit{double negation introduction}, and \textit{De Morgan's laws} (details in Tab.~\ref{Tab:rule_demo}). Each transformed formal representation is then rendered back into natural language using the same procedure as in the base dataset construction.
 This process yields three transformed datasets, \(\mathbb{D}_{\text{MI}}\), \(\mathbb{D}_{\text{DNI}}\), and \(\mathbb{D}_{\text{DM}}\), corresponding to material implication,  double negation introduction, and De Morgan's laws, respectively. Each transformed dataset is paired example-wise with \(\mathbb{D}_{\text{base}}\). We further construct a mixed dataset, \(\mathbb{D}_{\text{Mixed}}\), by randomly sampling a replacement rule at each reasoning step. Data examples are provided in App.~\ref{appendix:data_example}.

\vspace{-5pt}
\subsection{Evaluating LLMs on Logically Equivalent Data}
{We evaluate LLMs' capability for logic-invariant reasoning using our curated dataset, and identify two key observations: \textbf{(i)} LLMs fail to reliably capture logically equivalent structures, and \textbf{(ii)} prompting with explicit rules does not mitigate this failure. Specifically, we test models on $\mathbb{D}_{\text{base}}$
%\yy{are you sure it's base not mixed?} 
and its logically equivalent transformation \(\mathbb{D}_{\text{Mixed}}\) using two metrics: \textbf{Accuracy}, which measures correctness, and \textbf{Logic Inconsistency}, which measures the proportion of instances whose predicted label changes between the original input and its logically equivalent reformulation.
}

\begin{figure}[t]
    \centering
    \begin{subfigure}[t]{0.48\textwidth}
        \centering
        \includegraphics[width=\linewidth]{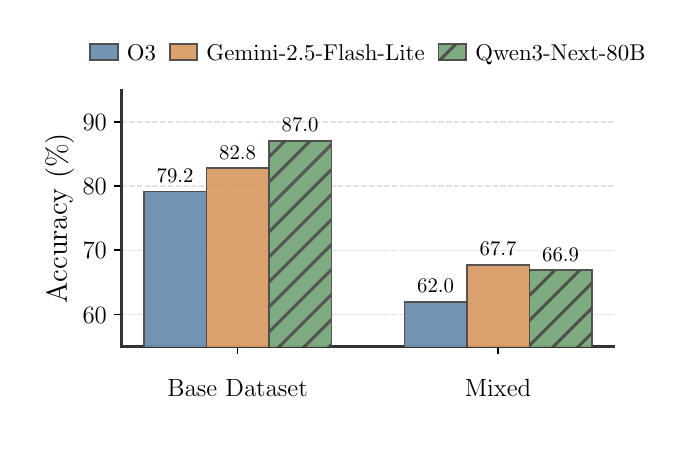}
         \vspace{-15pt}
        \caption{\small{Accuracy across logically equivalent datasets.  }}
        \label{fig:accuracy_analysis}
    \end{subfigure}
    \hfill
    \begin{subfigure}[t]{0.48\textwidth}
        \centering
        \includegraphics[width=\linewidth]{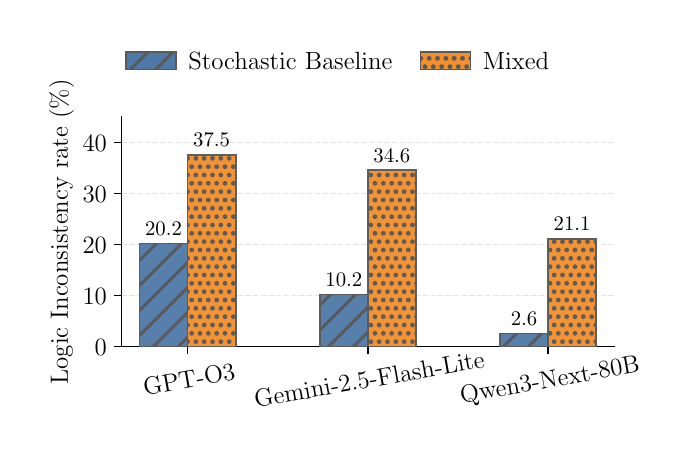}
         \vspace{-15pt}
        \caption{\small{Logic Inconsistency rate across logically equivalent datasets.}}
        \label{fig:fliprate_analysis}
    \end{subfigure}
     
    \caption{\small{Accuracy (\subref{fig:accuracy_analysis}) and Logical Inconsistency rates (\subref{fig:fliprate_analysis}) of LLMs under logically equivalent transformations. Even advanced models exhibit substantial performance drops and instability on the transformed inputs, revealing a fundamental limitation in capturing the underlying logical structure.}}
    \label{fig:two_subfigures}
    \vspace{-15pt}

\end{figure}

\paragraph{Observation 1: LLMs fail to capture logically equivalent structures.}
We evaluate closed-source models GPT-o3~\citep{openai2025o3o4mini} and Gemini-2.5-flash-lite~\citep{DBLP:journals/corr/abs-2507-06261}, along with Qwen3-Next-80B~\citep{DBLP:journals/corr/abs-2505-09388}, under a consistent protocol. 
{Results are shown in Fig.~\ref{fig:accuracy_analysis}. Although these models perform well in the original setting, their accuracy consistently drops in the mixed setting, where reasoning requires integrating equivalent reformulations across multiple steps. The drop reaches up to 20 percentage points, suggesting that the models struggle to reason consistently in logically equivalent forms.}
We next evaluate logical inconsistency. On $\mathbb{D}_{\text{Mixed}}$, the evaluated models show substantial disagreement between original inputs and their logically equivalent variants, with inconsistency rates between 21.1\% and 37.5\%. To contextualize this disagreement, we compare it against a \text{Stochastic Baseline}, which measures run-to-run disagreement on the original evaluation set and estimates variation due to decoding randomness alone. Reformulation-induced disagreement exceeds this baseline by approximately 20 percentage points across all models, indicating that the inconsistency primarily reflects sensitivity to logically equivalent reformulations rather than stochastic decoding variation.
\vspace{5pt}

\vspace{-15pt}

\begin{wraptable}{r}{0.44\textwidth}
\vspace{-1.0em}
\centering
\small
\caption{\small {Mixed reasoning performance (by \%): explicitly prompting LLMs to consider logical equivalence yields little to no improvement.}}
\label{tab:ICL_compare}
\renewcommand{\arraystretch}{1.03}
\setlength{\tabcolsep}{0.5pt}
\begin{tabular}{lc}
\toprule
\textbf{Model} & \textbf{Mixed } \\
\midrule
\texttt{\small{Gemini-2.5-flash-lite}}              & 67.7 \\
\texttt{\small{Gemini-2.5-flash-lite(w.Prompt)} }      & 70.4   \\ \midrule
\texttt{\small{Qwen3-Next-80B}}     &  66.9   \\
\texttt{Qwen3-Next-80B(w.Prompt)} & 66.1   \\
\bottomrule
\end{tabular}
\vspace{-1.0em}
\end{wraptable}

\vspace{5pt}
\paragraph{Observation 2: Prompting Is Insufficient to Induce Logic-invariant Reasoning.}

\label{para:icl}

We further examine whether explicit prompting improves LLM performance on logically equivalent data. In this setup, each model is prompted with a set of replacement rules, presented through natural-language explanations and corresponding formulas, as shown in Fig.~\ref{fig:icl_example}. We evaluate the model under the same protocol as above, with results reported in Tab.~\ref{tab:ICL_compare}. In-context prompting yields inconsistent and limited gains in the mixed reasoning setting, leaving a substantial performance gap from the original dataset. This indicates that merely presenting replacement rules in the prompt is insufficient for LLMs to reliably recognize or apply logical equivalence.

\paragraph{Logic structure as primary factor.} Finally, we extend our study to illustrate that the observed failures and inconsistency reflect an inability to recognize logical equivalence rather than other factors, \textit{e.g.}, differences in task difficulty after logic transformation. To do this, we first compare the \texttt{Mixed} variant with other logically equivalent variants that also involve complex rewrites, and we find prediction disagreement of $19.2\%$ between \texttt{Mixed} and \texttt{Mat.Impl.} and $18.3\%$ between \texttt{Mixed} and \texttt{Double Neg}. Both significantly exceed Qwen3-Next-80B’s stochastic noise level of $2.6\%$, showing that this trend persists even for data of comparable difficulty. The flip-direction analysis further supports this interpretation: if increased difficulty introduced by logical transformations drove these flips, correct-to-incorrect transitions would dominate as difficulty increased. However, in practice, we find that the transitions are nearly symmetric (GPT-o3, \texttt{Mixed} vs. \texttt{Mat.Impl.}: $12.5\%$ correct-to-incorrect versus $10.8\%$ incorrect-to-correct), with similar trends observed for other models. These findings further suggest that difficulty is not the primary driver of decision flips. 
Together, this evidence verifies that the primary reason for these inconsistencies is that LLMs struggle to recognize logical equivalence.

\section{Method: Logical Structure-guided Reasoning }
Building on our findings that current LLMs struggle with logic-invariant reasoning and that prompting alone does not reliably mitigate this issue, we turn to training-based methods, proposing LoGIST, which incorporates formal logical structure as auxiliary information to guide LLM reasoning. 

\begin{figure}
    \centering
    \includegraphics[width=0.95\linewidth]{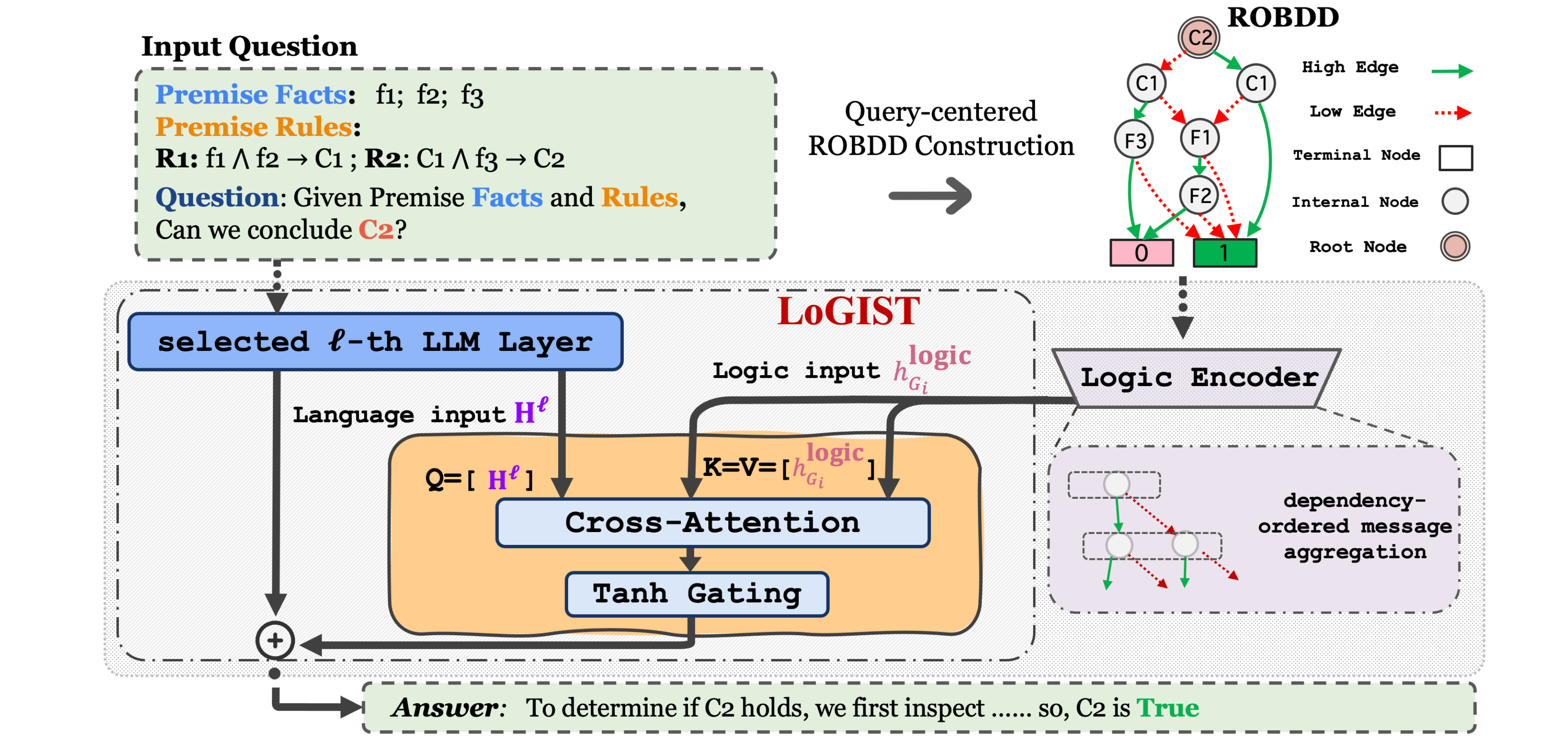}
    \caption{\small{
    Framework of LoGIST. LoGIST maps each question to an ROBDD that captures its logical dependency structure, then encodes the ROBDD into a logic embedding that conditions selected LLM layers via gated cross-attention. This lets the LLM attend to explicit logical-structure information while reasoning.}
    %\yy{Query-centered ROBDD construction sounds weird to me. Maybe use "ROBDD of the query"?}
    }
    \label{fig:logistframework}
    \vspace{-6pt}
\end{figure}

\subsection{Representing Logical Structure with ROBDDs}

To capture logical equivalence, we augment each natural-language input with an explicit logical-structure representation.
This representation should satisfy two properties: \textbf{(i)} it is grounded in truth-values, so that logically equivalent expressions share the same representation, and \textbf{(ii)} it captures the dependencies between grounded facts and rules. ROBDDs~\cite{DBLP:journals/tc/Bryant86,DBLP:journals/csur/Bryant92} satisfy both: under finite grounding and a fixed variable ordering, logically equivalent formulations over the same grounded atoms induce the same ROBDD, and the variable ordering itself encodes the reasoning dependencies. To construct the ROBDD for an instance, we determine its variable ordering by starting from the target query and recursively collecting the facts and rules required to evaluate its truth value. We then build the diagram using this ordering. We present the detailed procedure for this step in Alg.~\ref{alg:robdd_construction}.

\subsection{Learning Logical Embeddings}
\label{subsec:encoder_train}

Based on the constructed logic graph, our logic encoder aims to map each ROBDD to a graph-level representation while preserving its decision structure. Since an ROBDD represents a Boolean function through root-to-terminal decisions in a fixed order, the encoder in LoGIST should capture the ordered topology of internal nodes, distinguish between low and high branches, and preserve the Boolean values of terminal nodes. To this end, we design the encoder with partial-order-aware node updates, edge-specific propagation for the two branch types, and a terminal-based readout that yields the final graph-level representation.

{
Concretely, for an ROBDD $G_i$, each internal node is initialized by averaging the token embeddings of its associated facts. Node states are then updated in topological order so that each node is processed after all its predecessors have been updated. Let $\boldsymbol{h}_v$ denote the current hidden state of node $v$. Messages arriving through low and high edges are aggregated separately. For edge type $k \in \{0,1\}$, corresponding to low and high branches, respectively, the aggregated incoming message is $m_k(v) =
\sum_{(u,v) \in E_k(G_i)}
\sigma(\mathbf{W}_g^k \boldsymbol{h}_u)
\odot
(\mathbf{W}_m^k \boldsymbol{h}_u),$
where $E_k(G_i)$ denotes the set of edges of type $k$, and $\mathbf{W}_g^k$ and $\mathbf{W}_m^k$ are trainable edge-specific projections. {Here, $\sigma(\cdot)$ is the sigmoid function, and $\odot$ denotes element-wise multiplication. The sigmoid term acts as a dimension-wise gate that controls how much of the projected message $\mathbf{W}_m^k \boldsymbol{h}_u$ is passed from $u$ to $v$.}
The two aggregated messages are then summed and combined with the current state through a GRU: $\boldsymbol{h}'_v =
\mathrm{GRU}(m_0(v) + m_1(v), \boldsymbol{h}_v)$, which is used for subsequent propagation.
This propagation continues until the terminal nodes are updated, and the graph-level summary is read out from the two terminal states: $\mathbf{h}_{G_i}
=
\mathrm{MLP}([\boldsymbol{h}_{\tau_0} \| \boldsymbol{h}_{\tau_1}])
\in \mathbb{R}^{d_g}.$
The full encoding procedure is detailed in Alg.~\ref{alg:robdd_encoder}.}

Given this architecture, we train the encoder to learn representations that reflect ROBDD decision structure. To this end, we use contrastive learning~\citep{DBLP:conf/icml/0001I20,DBLP:conf/iclr/RobinsonCSJ21}, which pulls structurally equivalent ROBDDs closer together in the embedding space while pushing structurally distinct ones apart. To construct these pairs, we assign each ROBDD a structure-only signature using bottom-up hashing~\citep{10.1145/123186.123222,DBLP:conf/stoc/BabaiL83,DBLP:conf/stoc/Lindell92}. Each node is assigned a hash value based on the hash values of its children, with the root's hash value serving as the signature of the entire diagram. This signature depends solely on structural information, regardless of the original semantic content of the associated facts. ROBDDs sharing the same signature form positive pairs, while those with different signatures but similar sizes form hard negative pairs. The encoder is trained on these pairs using the InfoNCE loss~\citep{DBLP:journals/corr/abs-1807-03748}.

\subsection{Guiding LLM Reasoning with Logical Structure}

{While the contrastively trained encoder produces structure-aware embeddings, these embeddings lie in a representation space separate from that of the LLM, and directly incorporating them into the LLM's hidden representations can be brittle. We therefore use a two-stage strategy: \textit{logic-text alignment} maps logical embeddings into the LLM's representation space, and \textit{logic fusion} uses the aligned embeddings as auxiliary features to guide reasoning.}

{To align logical embeddings with the LLM representation space, we train a lightweight aligner $f_{\mathrm{align}}$ while keeping both the logic encoder and the LLM frozen. Given the graph embedding $\mathbf{h}_{G_i}$, the aligner projects it into the LLM hidden dimension: $\mathbf{h}^{\mathrm{logic}}_{G_i} = f_{\mathrm{align}}(\mathbf{h}_{G_i}) \in \mathbb{R}^{d_t}$, where $d_t$ is the LLM hidden dimension. We use the frozen LLM to obtain a paired text embedding $\mathbf{t}_{G_i}\in\mathbb{R}^{d_t}$ for each input by mean-pooling its hidden states. We also compute the mean $\boldsymbol{\mu}$ and top-$r$ PCA directions $\mathbf{U}\in\mathbb{R}^{r\times d_t}$ from all training text embeddings to characterize the dominant text-representation subspace.
The aligner is optimized with two complementary losses. A global subspace loss keeps aligned logical embeddings within the dominant LLM text-embedding subspace:}
\[
\mathcal{L}_{\mathrm{global}}(G_i)
=
\left\|
(\mathbf{h}^{\mathrm{logic}}_{G_i}-\boldsymbol{\mu})
-
\mathbf{U}^{\top}\mathbf{U}(\mathbf{h}^{\mathrm{logic}}_{G_i}-\boldsymbol{\mu})
\right\|_2^2,
\]
{An instance-level anchoring loss pulls each logical embedding toward its paired text embedding:}
\[
\mathcal{L}_{\mathrm{instance}}(G_i)
=
1-\cos\!\left(\mathbf{h}^{\mathrm{logic}}_{G_i},\mathbf{t}_{G_i}\right).
\]
{The final objective is: $\mathcal{L}_{\mathrm{align}}(G_i) = \mathcal{L}_{\mathrm{global}}(G_i)+\mathcal{L}_{\mathrm{instance}}(G_i)$.}
{To incorporate logical structure into the LLM, LoGIST treats the aligned logic embedding as an auxiliary conditioning signal. Rather than directly adding it to the token representations, which may disrupt the LLM’s original text representations, we use a cross-attention-based logic fusion module~\citep{DBLP:conf/nips/AlayracDLMBHLMM22}. At selected decoder layers, the textual hidden states attend to the aligned logic embedding, and the resulting logic-conditioned update is added back through a residual connection scaled by a learnable gate. This allows the model to selectively use logical information while preserving the original token representations.}
{
Specifically, at decoder layer $\ell$, let $\mathbf{H}_i^\ell \in \mathbb{R}^{T \times d_t}$ 
% denote the textual hidden states, 
{denote the textual hidden states for input $i$, where $T$ is the sequence length. The textual hidden states are used as queries, while the aligned logic embedding $\mathbf{h}^{\mathrm{logic}}_{G_i}$ provides the keys and values:}
% with $T$ the sequence length. The textual states serve as queries and the aligned logic embedding $\mathbf{h}^{\mathrm{logic}}_{G_i}$ as the key-value source:
$$
\mathbf{Q}_i^\ell = \mathbf{W}_Q^\ell \, \mathrm{LayerNorm}(\mathbf{H}_i^\ell), \qquad
\mathbf{K}_i^\ell = \mathbf{W}_K^\ell \, \mathbf{h}^{\mathrm{logic}}_{G_i}, \qquad
\mathbf{V}_i^\ell = \mathbf{W}_V^\ell \, \mathbf{h}^{\mathrm{logic}}_{G_i},
$$
where $\mathbf{W}_Q^\ell, \mathbf{W}_K^\ell, \mathbf{W}_V^\ell$ are learned projections. 
{The attention output is then projected and added to the original textual states through a gated residual update:}
$$
\widetilde{\mathbf{H}}_i^\ell
=
\mathbf{H}_i^\ell
+
\mathrm{tanh}(\alpha_\ell) \, \mathbf{W}_O^\ell \,
\mathrm{Attn}(\mathbf{Q}_i^\ell, \mathbf{K}_i^\ell, \mathbf{V}_i^\ell),
$$
where $\mathbf{W}_O^\ell$ is a learned output projection and $\alpha_\ell$ is
{a learnable scalar gate for layer $\ell$, which controls the strength of the logic-conditioned update.}
}

\section{Experiment}

In this section, we describe the basic experimental setup, including the dataset and evaluation metrics used. We then investigate three key research questions:  \textbf{RQ1}: Does LoGIST produce consistent predictions across logically equivalent inputs?
\textbf{RQ2}: Can LoGIST improve out-of-distribution generalization in logical reasoning?
\textbf{RQ3}: Can LoGIST serve as an effective logic encoding strategy for incorporating logical information into LLMs?

\subsection{Experimental Setup}

\paragraph{Datasets.}

We evaluate our method in two complementary settings. First, we test structural generalization under logical equivalence using a synthetic reasoning dataset. The training set contains 1{,}500 instances built from implicational rules.  We next generate another 900 examples using the same procedure to form a base evaluation set, and then apply the transformations in Sec.~\ref{subsec:rule_based} to obtain four out-of-distribution reformulations that are logically equivalent to the base examples.  Second, we conduct additional training and evaluation on real-world benchmarks, including FOLIO~\citep{DBLP:conf/emnlp/HanS0QRZCPQBSWS24} and ProverQA~\citep{qi2025large}, to assess effectiveness beyond our controlled setting. For example, human experts wrote FOLIO. For FOLIO, we follow the standard split. For ProverQA, we use the split that provides paired formal representations, partitioned into 1{,}000 training and 500 evaluation instances.

\paragraph{Training Details.} To evaluate out-of-distribution (OOD) generalization to unseen logical forms, we adopt a strict training protocol. We first train the logic encoder solely on ROBDDs from the synthetic implicational training set, then freeze it and jointly train the LLM and the logic fusion module. This frozen encoder is shared by all experiments below. On the synthetic dataset, the LLM is trained only on the implicational training set and evaluated on the base evaluation set and four transformed variants, testing if the learned structure generalizes to unseen logical forms. For FOLIO and ProverQA, we keep the same synthetic-trained logic encoder frozen, train the LLM on each dataset’s training set, and evaluate on its corresponding test set. We apply LoGIST to \texttt{Llama-3.1-8B-Ins}~\citep{DBLP:journals/corr/abs-2407-21783} and \texttt{Qwen2.5-3B-Ins}~\citep{DBLP:journals/corr/abs-2412-15115} under three training paradigms: SFT with CoT supervision, PPO~\citep{DBLP:journals/corr/SchulmanWDRK17}, and GRPO~\citep{DBLP:journals/corr/abs-2402-03300}. In each setting, we compare a baseline against its LoGIST-augmented counterpart under identical configurations. Further training details are provided in App.~\ref{appendix:training_config}.

\subsection{RQ1: LoGIST Improves Consistency Across Logically Equivalent Inputs}

\label{subsec:RQ1}
\begin{wraptable}{r}{0.52\textwidth}
\vspace{-0.3cm}
\caption{\small Logic Inconsistency rate across logically equivalent transformations on \texttt{Llama-3.1-8B-Ins}. A lower score indicates more coherent reasoning.}
\label{tab:inconsistency_rate}
\vspace{-0.4em}
\centering
\footnotesize
\setlength{\tabcolsep}{3.6pt}
\renewcommand{\arraystretch}{1.15}
\resizebox{\linewidth}{!}{
\begin{tabular}{lcccc}
\toprule
\textbf{Methods} & \textbf{Mat.Impl.} & \textbf{Double Neg.} & \textbf{De Morgan} & \textbf{Mixed} \\
\midrule
SFT        & 25.7\% & 18.4\% & 17.9\% & 19.0\% \\
SFT(LoGIST)   & \textbf{20.1}\% & \textbf{14.2}\% & \textbf{13.1}\%& \textbf{15.8}\% \\
\midrule
PPO         & 31.5\% & 29.7\% & 32.0\% & 30.6\% \\
PPO(LoGIST)    & \textbf{15.3}\% & \textbf{15.9}\% & \textbf{15.9}\% & \textbf{13.9}\% \\
\midrule
GRPO        & 38.1\% & 32.5\% & 34.7\% & 37.1\% \\
GRPO(LoGIST)   & \textbf{19.9}\% & \textbf{20.1}\% & \textbf{22.4}\% & \textbf{16.6}\% \\
\bottomrule
\end{tabular}
}
\vspace{-1.0em}
\end{wraptable}

We first evaluate whether LoGIST stabilizes LLM predictions across logically equivalent variants using the synthetic dataset. Following the setup described above, we train the LLMs only on the implicational training set, and then compute the logic inconsistency rate by comparing predictions on the base set with those on each transformed set, averaging over three runs per dataset. As shown in Tab.~\ref{tab:inconsistency_rate}, LoGIST nearly halves the average inconsistency rate compared with standard training, reducing it from above 0.3 to roughly 0.15 under both PPO and GRPO, with consistent gains under SFT. The reduction holds across all four transformations, indicating that it is not driven by any single type of variation. This suggests that LoGIST improves prediction consistency through logical structure rather than surface patterns, enabling generalization to logical forms unseen during training.

\subsection{RQ2:  LoGIST Improves Generalization in Logical Reasoning}

\begin{table*}[ht]
\centering
\caption{\small{Performance on the evaluation set for \texttt{Llama-3.1-8B-Ins} and \texttt{Qwen2.5-3B-Ins}, with and without LoGIST, across different training strategies.  ``/'' indicates unavailable SFT results on FOLIO due to missing CoT labels. Colored cells indicate better performance.  LoGIST consistently improves performance across settings.}}
\label{table:main_res}
\renewcommand{\arraystretch}{1.08}
\resizebox{\textwidth}{!}{%
\begin{tabular}{c|l|c|cccc|c|c}
\toprule
\multirow{2}{*}{\small\textbf{LLM}} 
& \multirow{2}{*}{\textbf{Methods}}
& \multicolumn{5}{c|}{\small\textbf{Synthetic}} 
& \multirow{2}{*}{\small\textbf{FOLIO}}
& \multirow{2}{*}{\small\textbf{ProverQA}} \\
\cmidrule(lr){3-7}
& 
& \small\textbf{Base}
& \small\textbf{Mat.Impl.}
& \small\textbf{Double Neg.}
& \small\textbf{De Morgan}
& \small\textbf{Mixed}
& & \\
\midrule

\multirow{6}{*}{\rotatebox{90}{\texttt{Llama3.1-8B-ins}}}
& + SFT
& 0.683 & 0.588 & 0.625 & 0.631 & 0.635 & / & 0.598 \\ 
& + SFT(\small{LoGIST})
& \cellcolor[HTML]{fcf1e2}0.693
& \cellcolor[HTML]{fcf1e2}0.604
& \cellcolor[HTML]{fcf1e2}0.650
& \cellcolor[HTML]{fcf1e2}0.647
& \cellcolor[HTML]{fcf1e2}0.651
& /
& \cellcolor[HTML]{fcf1e2}0.638 \\ 
\cmidrule(lr){2-9}
& + PPO
& \cellcolor[HTML]{fcf1e2}0.710 & 0.614 & 0.621 & 0.590 & 0.618 & 0.606 & 0.568 \\
& + PPO(\small{LoGIST})
& 0.708
& \cellcolor[HTML]{fcf1e2}0.655
& \cellcolor[HTML]{fcf1e2}0.657
& \cellcolor[HTML]{fcf1e2}0.645
& \cellcolor[HTML]{fcf1e2}0.642
& \cellcolor[HTML]{fcf1e2}0.645
& \cellcolor[HTML]{fcf1e2}0.602 \\ 
\cmidrule(lr){2-9}
& + GRPO
& 0.702 & 0.608 & 0.612 & 0.600 & 0.627 & 0.621 & 0.626 \\
& + GRPO(\small{LoGIST})
& \cellcolor[HTML]{fcf1e2}0.730
& \cellcolor[HTML]{fcf1e2}0.654
& \cellcolor[HTML]{fcf1e2}0.661
& \cellcolor[HTML]{fcf1e2}0.647
& \cellcolor[HTML]{fcf1e2}0.668
& \cellcolor[HTML]{fcf1e2}0.650
& \cellcolor[HTML]{fcf1e2}0.652 \\

\midrule
\midrule

\multirow{6}{*}{\rotatebox{90}{\texttt{Qwen2.5-3B-Ins}}}
& + SFT
& 0.662 & 0.579 & 0.610 & 0.601 & 0.604 & / & 0.582 \\ 
& + SFT(\small{LoGIST})
& \cellcolor[HTML]{fcf1e2}0.674
& \cellcolor[HTML]{fcf1e2}0.607
& \cellcolor[HTML]{fcf1e2}0.638
& \cellcolor[HTML]{fcf1e2}0.616
& \cellcolor[HTML]{fcf1e2}0.630
& /
& \cellcolor[HTML]{fcf1e2}0.624 \\ 
\cmidrule(lr){2-9}
& + PPO
&  \cellcolor[HTML]{fcf1e2} 0.690 & 0.631 & 0.645 & 0.618 & 0.636 & 0.596 & 0.594 \\
& + PPO(\small{LoGIST})
& 0.681
& \cellcolor[HTML]{fcf1e2}0.661
& \cellcolor[HTML]{fcf1e2}0.658
& \cellcolor[HTML]{fcf1e2}0.658
& \cellcolor[HTML]{fcf1e2}0.657
& \cellcolor[HTML]{fcf1e2}0.606
& \cellcolor[HTML]{fcf1e2}0.608 \\ 
\cmidrule(lr){2-9}
& + GRPO
& 0.645 & 0.570 & 0.577 & 0.572 & 0.590 & 0.621 & 0.600 \\
& + GRPO(\small{LoGIST})
& \cellcolor[HTML]{fcf1e2}0.655
& \cellcolor[HTML]{fcf1e2}0.604
& \cellcolor[HTML]{fcf1e2}0.607
& \cellcolor[HTML]{fcf1e2}0.598
& \cellcolor[HTML]{fcf1e2}0.620
& \cellcolor[HTML]{fcf1e2}0.631
& \cellcolor[HTML]{fcf1e2}0.616 \\

\bottomrule
\end{tabular}%
}
\vspace{-0.2em}
\end{table*}

We next examine whether improved logical invariance leads to stronger generalization. We evaluate accuracy on both synthetic and real-world datasets. As shown in Tab.~\ref{table:main_res}, LoGIST consistently improves performance across nearly all algorithms and datasets. On the synthetic dataset, the gains are largest on OOD logical forms, showing that structure learned from implicational training data transfers to logically equivalent forms unseen during training. We also find that the gains are more pronounced for RL-based methods: for example, on \texttt{Llama-3.1-8B-Ins}, LoGIST improves GRPO’s performance by $0.046$ on \texttt{Mat.Impl}. and $0.041$ on \texttt{Mixed}, suggesting that learning logical structure may be particularly beneficial when training relies on terminal rewards rather than step-level CoT supervision. On FOLIO and ProverQA, LoGIST also improves accuracy, indicating that the synthetic-trained logic encoder captures structural information useful for real-world reasoning tasks. Together, these results show that learning from logically equivalent structures improves LLM generalization, and that the consistency gains in Sec.~\ref{subsec:RQ1} translate into meaningful accuracy improvements rather than merely reducing prediction flips. We further show in App.~\ref{appendix:robust_pratical} that LoGIST remains effective even in noisy, practical settings where gold formal representations are unavailable.

\subsection{RQ3: LoGIST is an Effective Logic Encoding Strategy }

\begin{wraptable}{r}{0.56\textwidth}
\vspace{-0.4cm}
\begin{minipage}{\linewidth}
\centering
\caption{\small{Performance comparison across different strategies incorporating ROBDD information. LoGIST achieves better performance than using Linearized ROBDD.}}
\label{tab:sft_logist_results}

\renewcommand{\arraystretch}{1.15}
\resizebox{\linewidth}{!}{%
\begin{tabular}{lccccc}
\toprule
\textbf{Method} & \textbf{Base} & \textbf{M.I.} & \textbf{D.N.} &
\textbf{D.M.} & \textbf{Mixed} \\
\midrule
SFT
    & 0.683 & 0.588 & 0.625 & 0.631 & 0.635 \\
SFT + Linear ROBDD
    & 0.674 & 0.594 & 0.642 & 0.615 & 0.639 \\
SFT + LoGIST
    & \textbf{0.693} & \textbf{0.604} & \textbf{0.650}
    & \textbf{0.647} & \textbf{0.651} \\
\bottomrule
\end{tabular}%
}
\end{minipage}
\vspace{-0.4em}
\end{wraptable}

We further examine whether LoGIST's gains stem simply from providing the LLM with additional logical structure information in the form of an ROBDD, rather than from how this information is encoded and integrated. To test this, we introduce a baseline that receives the complete ROBDD as text. Following \cite{fatemi2024talk}, we linearize the diagram into a node-by-node description and include it in the prompt alongside the context (Fig.~\ref{fig:linearized_robdd}). This baseline therefore has access to the same ROBDD as LoGIST; the two approaches differ only in how the structure is provided: as prompt text versus as a learned logic embedding fused through cross-attention.

As shown in Tab.~\ref{tab:sft_logist_results}, providing the linearized ROBDD as prompt text yields only marginal and inconsistent changes in accuracy (\textit{e.g.}, $+0.4\%$ on \texttt{Mixed}, $-0.9\%$ on \texttt{Base}, and $-1.6\%$ on \texttt{D.M.}). Moreover, it does not reduce the inconsistency and instead increases it from $19.0\%$ to $21.3\%$ in Mixed. We further repeat this comparison under PPO and GRPO, where the performance gap becomes even larger. With only terminal rewards available, the LLM struggles to learn how to effectively utilize the verbalized ROBDD information provided in the prompt. These results demonstrate that LoGIST’s performance gains stem from its effective utilization of logical structure, rather than merely from access to additional logical information.

\subsection{Analysis: Composition as a Linear Direction }
\label{sec:linear_direction}

Having shown the effectiveness of our trained logic encoder, we next examine how logical structure is reflected in its representation space. Specifically, we examine how the composition of consecutive reasoning steps is represented in this space by comparing the resulting logic embedding with the sum of their individual embeddings. We ask whether this composition follows a consistent geometric pattern across instances. Motivated by prior work linking linear structure in representation spaces to generalization~\citep{DBLP:conf/nips/YanSKRH20,DBLP:conf/iccv/TragerPZABS23,DBLP:conf/iclr/ParkLLYONWT25,DBLP:conf/nips/HeDDG24,DBLP:conf/acl/LiuHYY25}, we test whether the composed embedding can be approximated by this sum plus a shared linear shift in the learned logical embedding space.

To test this hypothesis, we measure whether adding a reasoning step behaves approximately linearly in the learned embedding space. For each instance
$i$ with $N$ reasoning steps, we construct three ROBDDs {at step $n$}: $G_i^{n}$, {which contains only the $n$-th rule};
$G_i^{<n}$, 
{which contains the rules preceding step $n$;}
and $G_i^{\le n}$,
{which contains all rules up to and including step $n$.}
Let $\mathbf{h}_{G_i,n}^{\mathrm{logic}}$, $\mathbf{h}_{G_i,<n}^{\mathrm{logic}}$, and $\mathbf{h}_{G_i,\le n}^{\mathrm{logic}}$ denote the {corresponding logic embeddings}. We define the compositional residual at step $n$ as: $
\boldsymbol{\delta}_i^{\,n}
=
\mathbf{h}_{G_i,\le n}^{\mathrm{logic}}
-
\bigl(\mathbf{h}_{G_i,<n}^{\mathrm{logic}}+\mathbf{h}_{G_i,n}^{\mathrm{logic}}\bigr).
$ 
This residual captures the part of the composed representation that is not explained by simply adding the representation of the preceding rules and that of the isolated $n$-th rule. If the same reasoning step follows a shared compositional pattern across instances, then the residuals
$\boldsymbol{\delta}_i^{\,n}$ should align along a common direction rather than vary independently.

{To test whether the residuals indeed concentrate on a shared direction, we adopt the leave-one-out (LOO) alignment score~\citep{DBLP:conf/icml/ParkCV24}: for each instance, we check whether its residual aligns with the mean direction estimated from all other instances. Specifically, we first whiten each residual to reduce representation anisotropy:  $
\tilde{\boldsymbol{\delta}}_i^{\,n}
=
\mathbf{\Omega}\bigl(\boldsymbol{\delta}_i^{\,n}-\boldsymbol{\hat{\mu}}\bigr),
$
where $\boldsymbol{\hat{\mu}}$ is the mean of the pooled residuals and $\mathbf{\Omega}$ is a whitening operator. We then normalize each whitened residual as $
\hat{\tilde{\boldsymbol{\delta}}}_i^{\,n}
=
\frac{\tilde{\boldsymbol{\delta}}_i^{\,n}}
{\|\tilde{\boldsymbol{\delta}}_i^{\,n}\|_2},
$ and compute the LOO mean direction by averaging over all other instances as $
\mathbf{u}_{-i}^{\,n}
=
\frac{\sum_{j\neq i}\hat{\tilde{\boldsymbol{\delta}}}_j^{\,n}}
{\left\|\sum_{j\neq i}\hat{\tilde{\boldsymbol{\delta}}}_j^{\,n}\right\|_2}.
$ The LOO alignment score is then the inner product $
\left\langle
\hat{\tilde{\boldsymbol{\delta}}}_i^{\,n},
\mathbf{u}_{-i}^{\,n}
\right\rangle,
$ which is high when the residual of instance $i$ is well explained by the shared direction induced by the rest.
}

\begin{wrapfigure}{r}{0.45\linewidth}
\centering
 \vspace{-0.3cm}
\includegraphics[width=\linewidth]{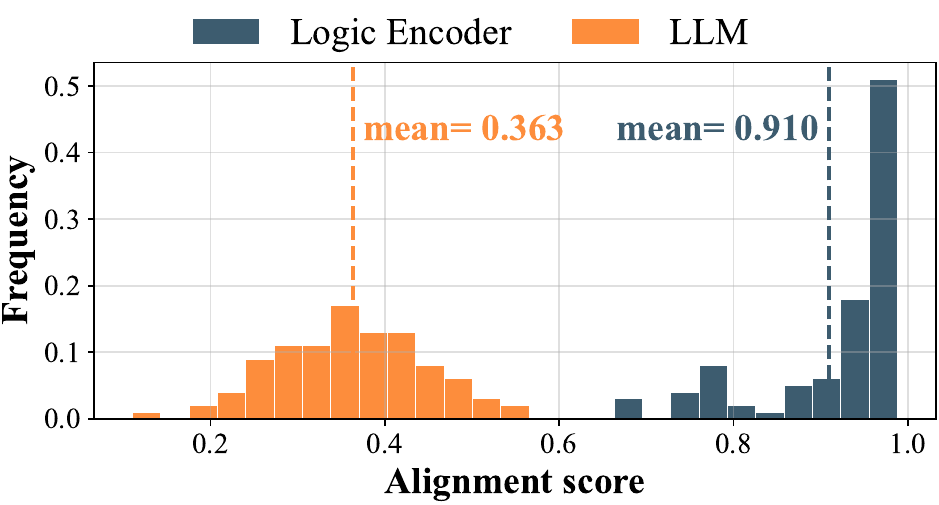}
\vspace{-0.3cm}
\caption{\small Distributional comparison of the per-instance LOO alignment scores for \texttt{Llama-3.1-8B-Ins} and its logic encoder.}
    \label{fig:align_score}
    \vspace{-0.5cm}
\end{wrapfigure}

To verify that this linear compositional structure is specific to our logic encoder rather than a generic property of LLM representations, we compare LOO alignment scores between two representations: the aligned logic embedding $\mathbf{h}^{\mathrm{logic}}_{G_i}$, and the mean-pooled final-layer LLM hidden state of the corresponding natural-language input. Although the two vectors come from different sources, they share the same dimensionality and can thus be directly compared.

As shown in Fig.~\ref{fig:align_score}, the logic encoder yields high per-instance LOO alignment scores concentrated near $0.9$ (mean $= 0.910$), whereas the LLM text representations score much lower (mean $=0.363$).  This suggests that the shared residual direction is learned by the logic encoder rather than already present in the LLM representation space. This direction is also stable across reasoning depths: the LOO mean directions estimated at different reasoning steps have an average pairwise cosine similarity of 0.90, indicating that the same compositional pattern is reused from one step to the next.  Together, these results suggest that the logic encoder captures reasoning-step composition through a shared linear structure, providing evidence for its ability to support generalization across logical forms.

We further include a study of LoGIST’s efficiency in App.~\ref{appendix:efficiency}, where we show that LoGIST introduces only minimal overhead and negligible additional latency. We also demonstrate LoGIST’s robustness in App.~\ref{appendix:robust_pratical}, showing that it maintains its performance even when ground-truth formal logic representations are unavailable.
\vspace{-4pt}
\section{Related Work}
\vspace{-4pt}

\paragraph{LLMs for Formal Reasoning}

Formal reasoning systems such as Lean~\citep{DeMoura2015Lean}, Z3~\citep{10.5555/1792734.1792766}, and Prover9~\citep{McCune2005Prover9} have long provided verifiable inference with guaranteed soundness. Recent work extends LLMs to formal reasoning along two directions. Data-driven methods improve LLMs' reasoning capability through curated training data ~\citep{DBLP:conf/emnlp/HanS0QRZCPQBSWS24,DBLP:conf/acl/TafjordDC21,DBLP:conf/nips/SaparovPPJKK023}, or through synthesized logical reasoning data~\citep{qi2025large}. These approaches optimize task-level accuracy without modeling logical structure, limiting generalization on unseen logical forms. External-tool methods couple LLMs with symbolic solvers to improve reasoning reliability~\citep{pan-etal-2023-logic,DBLP:conf/nips/YeCDD23,olausson2023linc}, with recent work further enhancing solver use through semantic verification across alternative translations~\citep{DBLP:conf/iclr/RyuKLY25} or sub-proposition feedback for iterative proof correction~\citep{DBLP:conf/emnlp/CaoLDYZZSLHG25}. While effective, these methods rely on external systems for formal reasoning, leaving the LLM’s own structure-aware reasoning capability underexplored. In contrast, we show that learning from logical equivalence can improve generalizable logical reasoning in LLMs.

\paragraph{Multimodal LLMs} Extending LLMs beyond pure text to additional modalities has long been an active research field. \citep{DBLP:conf/icml/RadfordKHRGASAM21,DBLP:conf/iclr/Chen0CPPSGGMB0P23,DBLP:conf/icml/JiaYXCPPLSLD21} study large-scale vision-language pretraining and cross-modal alignment, while \citep{DBLP:conf/nips/LiuLWL23a,DBLP:conf/nips/AlayracDLMBHLMM22} further connect visual features with pretrained language models for multimodal understanding. Beyond visual inputs, recent work has explored modalities related to symbolic information: \citep{DBLP:journals/corr/abs-2509-25851} benchmarks symbolic reasoning in LLMs using multimodal inputs; \citep{Lin2024GT2VecLL} uses graphs as an additional modality for joint graph-text representation learning; \citep{li2026table} encodes table schemas to enhance tabular reasoning; \cite{DBLP:conf/icml/GongEC24,DBLP:conf/nips/DingPMKYR24}  encode code structure and syntax as a modality to improve LLMs' understanding of code semantics; and \citep{pei2024exploiting} exploits code symmetries for code-LLM training. However, to our knowledge, our work is the first to model logical structure to guide logic-invariant reasoning in LLMs and further improve their generalizability.
\vspace{-4pt}
\section{Conclusion}
\vspace{-4pt}

{In this work, we identify a key limitation of current LLMs: strong reasoning performance does not necessarily imply a precise understanding of the underlying logical structure. Through systematic evaluation on logically equivalent reformulations, we show that even strong LLMs produce inconsistent predictions and exhibit degraded performance when the same logic is expressed in different forms, limiting their ability to generalize to unseen logical forms. To address this limitation, we propose LoGIST, a logic-structure-guided framework that maps each instance to a decision diagram, encodes it into a structure-aware logic embedding through contrastive learning, and conditions LLM reasoning on this embedding. Across different training paradigms and both synthetic and real-world benchmarks, LoGIST consistently reduces logical inconsistency and improves generalization to unseen logical forms. Further analysis shows that LoGIST encodes logical composition as a linear direction, which helps explain the observed gains in generalization. These results suggest that logical invariance is an effective target for building more robust and generalizable LLM reasoning systems.}

\vspace{-4pt}
\section*{Acknowledgments}
We thank the anonymous reviewers for their effort in helping us revise our study. This material is based upon work supported by the National Science Foundation under CAREER Grant No. IIS 2543142. 
This research was also supported in part by the National Science Foundation (NSF) under Grant No. DMS 2412577. 
Any opinions, findings, and conclusions, or recommendations expressed in this material are those of the author(s) and do not necessarily reflect the views of the National Science Foundation.
\vspace{-4pt}

\bibliographystyle{plain} % Defines your bibliography style
\bibliography{ref}

@inproceedings{
qi2025large,
title={Large Language Models Meet Symbolic Provers for Logical Reasoning Evaluation},
author={Chengwen Qi and Ren Ma and Bowen Li and He Du and Binyuan Hui and Jinwang Wu and Yuanjun Laili and Conghui He},
booktitle={The Thirteenth International Conference on Learning Representations},
year={2025},
url={https://openreview.net/forum?id=C25SgeXWjE}
}

@inproceedings{DBLP:conf/nips/DingPMKYR24,
  author       = {Yangruibo Ding and
                  Jinjun Peng and
                  Marcus J. Min and
                  Gail E. Kaiser and
                  Junfeng Yang and
                  Baishakhi Ray},
  editor       = {Amir Globersons and
                  Lester Mackey and
                  Danielle Belgrave and
                  Angela Fan and
                  Ulrich Paquet and
                  Jakub M. Tomczak and
                  Cheng Zhang},
  title        = {SemCoder: Training Code Language Models with Comprehensive Semantics
                  Reasoning},
  booktitle    = {Advances in Neural Information Processing Systems 37: Annual Conference
                  on Neural Information Processing Systems 2024, NeurIPS 2024, Vancouver,
                  BC, Canada, December 10 - 15, 2024},
  year         = {2024},
  url          = {http://papers.nips.cc/paper\_files/paper/2024/hash/6efcc7fd8efeee29a050a79c843c90e0-Abstract-Conference.html},
  bibsource    = {dblp computer science bibliography, https://dblp.org}
}

@inproceedings{DBLP:conf/icml/GongEC24,
  author       = {Linyuan Gong and
                  Mostafa Elhoushi and
                  Alvin Cheung},
  editor       = {Ruslan Salakhutdinov and
                  Zico Kolter and
                  Katherine A. Heller and
                  Adrian Weller and
                  Nuria Oliver and
                  Jonathan Scarlett and
                  Felix Berkenkamp},
  title        = {{AST-T5:} Structure-Aware Pretraining for Code Generation and Understanding},
  booktitle    = {Forty-first International Conference on Machine Learning, {ICML} 2024,
                  Vienna, Austria, July 21-27, 2024},
  series       = {Proceedings of Machine Learning Research},
  volume       = {235},
  pages        = {15839--15853},
  publisher    = {{PMLR} / OpenReview.net},
  year         = {2024},
  url          = {https://proceedings.mlr.press/v235/gong24c.html},
  bibsource    = {dblp computer science bibliography, https://dblp.org}
}

@article{DBLP:journals/tmlr/PoesiaGZG24,
  author       = {Gabriel Poesia and
                  Kanishk Gandhi and
                  Eric Zelikman and
                  Noah D. Goodman},
  title        = {Certified Deductive Reasoning with Language Models},
  journal      = {Trans. Mach. Learn. Res.},
  volume       = {2024},
  year         = {2024},
  url          = {https://openreview.net/forum?id=yXnwrs2Tl6},
  bibsource    = {dblp computer science bibliography, https://dblp.org}
}

@inproceedings{DBLP:conf/nips/ZhouP0CCLCZMZ24,
  author       = {Pei Zhou and
                  Jay Pujara and
                  Xiang Ren and
                  Xinyun Chen and
                  Heng{-}Tze Cheng and
                  Quoc V. Le and
                  Ed H. Chi and
                  Denny Zhou and
                  Swaroop Mishra and
                  Huaixiu Steven Zheng},
  editor       = {Amir Globersons and
                  Lester Mackey and
                  Danielle Belgrave and
                  Angela Fan and
                  Ulrich Paquet and
                  Jakub M. Tomczak and
                  Cheng Zhang},
  title        = {{SELF-DISCOVER:} Large Language Models Self-Compose Reasoning Structures},
  booktitle    = {Advances in Neural Information Processing Systems 37: Annual Conference
                  on Neural Information Processing Systems 2024, NeurIPS 2024, Vancouver,
                  BC, Canada, December 10 - 15, 2024},
  year         = {2024},
  url          = {http://papers.nips.cc/paper\_files/paper/2024/hash/e41efb03e20ca3c231940a3c6917ef6f-Abstract-Conference.html},
  bibsource    = {dblp computer science bibliography, https://dblp.org}
}

@inproceedings{DBLP:conf/icml/MorishitaMYS23,
  author       = {Terufumi Morishita and
                  Gaku Morio and
                  Atsuki Yamaguchi and
                  Yasuhiro Sogawa},
  editor       = {Andreas Krause and
                  Emma Brunskill and
                  Kyunghyun Cho and
                  Barbara Engelhardt and
                  Sivan Sabato and
                  Jonathan Scarlett},
  title        = {Learning Deductive Reasoning from Synthetic Corpus based on Formal
                  Logic},
  booktitle    = {International Conference on Machine Learning, {ICML} 2023, 23-29 July
                  2023, Honolulu, Hawaii, {USA}},
  series       = {Proceedings of Machine Learning Research},
  volume       = {202},
  pages        = {25254--25274},
  publisher    = {{PMLR}},
  year         = {2023},
  url          = {https://proceedings.mlr.press/v202/morishita23a.html},
  bibsource    = {dblp computer science bibliography, https://dblp.org}
}

@inproceedings{DBLP:conf/acl/LiuCDLZWYCC26,
  author       = {Xuyuan Liu and
                  Shengyu Chen and
                  Xinshuai Dong and
                  Yanchi Liu and
                  Xujiang Zhao and
                  Haoyu Wang and
                  Yujun Yan and
                  Haifeng Chen and
                  Zhengzhang Chen},
  editor       = {Maria Liakata and
                  Viviane P. Moreira and
                  Jiajun Zhang and
                  David Jurgens},
  title        = {Representation Interventions Enable Lifelong Knowledge Memory Control
                  in LLMs},
  booktitle    = {Proceedings of the 64th Annual Meeting of the Association for Computational
                  Linguistics (Volume 1: Long Papers), {ACL} 2026, San Diego, California,
                  United States, July 2-7, 2026},
  pages        = {5414--5436},
  publisher    = {Association for Computational Linguistics},
  year         = {2026},
  url          = {https://doi.org/10.18653/v1/2026.acl-long.246},
  doi          = {10.18653/V1/2026.ACL-LONG.246},
  bibsource    = {dblp computer science bibliography, https://dblp.org}
}

@inproceedings{DBLP:conf/acl/LiuHYY25,
  author       = {Xuyuan Liu and
                  Lei Hsiung and
                  Yaoqing Yang and
                  Yujun Yan},
  editor       = {Wanxiang Che and
                  Joyce Nabende and
                  Ekaterina Shutova and
                  Mohammad Taher Pilehvar},
  title        = {Spectral Insights into Data-Oblivious Critical Layers in Large Language
                  Models},
  booktitle    = {Findings of the Association for Computational Linguistics, {ACL} 2025,
                  Vienna, Austria, July 27 - August 1, 2025},
  series       = {Findings of {ACL}},
  volume       = {{ACL} 2025},
  pages        = {4860--4877},
  publisher    = {Association for Computational Linguistics},
  year         = {2025},
  url          = {https://doi.org/10.18653/v1/2025.findings-acl.251},
  doi          = {10.18653/V1/2025.FINDINGS-ACL.251},
  bibsource    = {dblp computer science bibliography, https://dblp.org}
}

@inproceedings{
fatemi2024talk,
title={Talk like a Graph: Encoding Graphs for Large Language Models},
author={Bahare Fatemi and Jonathan Halcrow and Bryan Perozzi},
booktitle={The Twelfth International Conference on Learning Representations},
year={2024},
url={https://openreview.net/forum?id=IuXR1CCrSi}
}

@inproceedings{putra-etal-2026-nl2logic,
    title = "{NL}2{L}ogic: {AST}-Guided Translation of Natural Language into First-Order Logic with Large Language Models",
    author = "Putra, Rizky Ramadhana  and
      Basuki, Raihan Sultan Pasha  and
      Cheng, Yutong  and
      Gao, Peng",
    editor = "Demberg, Vera  and
      Inui, Kentaro  and
      Marquez, Llu{\'i}s",
    booktitle = "Findings of the {A}ssociation for {C}omputational {L}inguistics: {EACL} 2026",
    month = mar,
    year = "2026",
    address = "Rabat, Morocco",
    publisher = "Association for Computational Linguistics",
    url = "https://aclanthology.org/2026.findings-eacl.317/",
    doi = "10.18653/v1/2026.findings-eacl.317",
    pages = "6035--6051",
    ISBN = "979-8-89176-386-9"
}

@book{10.5555/579402,
author = {Rosen, Kenneth H.},
title = {Discrete Mathematics and Its Applications},
year = {2002},
isbn = {0072424346},
publisher = {McGraw-Hill Higher Education},
edition = {5th}
}

@inproceedings{DBLP:conf/stoc/Lindell92,
  author       = {Steven Lindell},
  editor       = {S. Rao Kosaraju and
                  Mike Fellows and
                  Avi Wigderson and
                  John A. Ellis},
  title        = {A Logspace Algorithm for Tree Canonization (Extended Abstract)},
  booktitle    = {Proceedings of the 24th Annual {ACM} Symposium on Theory of Computing,
                  May 4-6, 1992, Victoria, British Columbia, Canada},
  pages        = {400--404},
  publisher    = {{ACM}},
  year         = {1992},
  url          = {https://doi.org/10.1145/129712.129750},
  doi          = {10.1145/129712.129750},
  bibsource    = {dblp computer science bibliography, https://dblp.org}
}

@inproceedings{DBLP:conf/acl/TafjordDC21,
  author       = {Oyvind Tafjord and
                  Bhavana Dalvi and
                  Peter Clark},
  editor       = {Chengqing Zong and
                  Fei Xia and
                  Wenjie Li and
                  Roberto Navigli},
  title        = {ProofWriter: Generating Implications, Proofs, and Abductive Statements
                  over Natural Language},
  booktitle    = {Findings of the Association for Computational Linguistics: {ACL/IJCNLP}
                  2021, Online Event, August 1-6, 2021},
  series       = {Findings of {ACL}},
  pages        = {3621--3634},
  publisher    = {Association for Computational Linguistics},
  year         = {2021},
  url          = {https://doi.org/10.18653/v1/2021.findings-acl.317},
  doi          = {10.18653/V1/2021.FINDINGS-ACL.317},
  bibsource    = {dblp computer science bibliography, https://dblp.org}
}

@inproceedings{DBLP:conf/iclr/RyuKLY25,
  author       = {Hyun Ryu and
                  Gyeongman Kim and
                  Hyemin S. Lee and
                  Eunho Yang},
  title        = {Divide and Translate: Compositional First-Order Logic Translation
                  and Verification for Complex Logical Reasoning},
  booktitle    = {The Thirteenth International Conference on Learning Representations,
                  {ICLR} 2025, Singapore, April 24-28, 2025},
  publisher    = {OpenReview.net},
  year         = {2025},
  url          = {https://openreview.net/forum?id=09FiNmvNMw},
  bibsource    = {dblp computer science bibliography, https://dblp.org}
}

@inproceedings{DBLP:conf/acl/DuSSWSC24,
  author       = {Kevin Du and
                  V{\'{e}}steinn Sn{\ae}bjarnarson and
                  Niklas Stoehr and
                  Jennifer C. White and
                  Aaron Schein and
                  Ryan Cotterell},
  editor       = {Lun{-}Wei Ku and
                  Andre Martins and
                  Vivek Srikumar},
  title        = {Context versus Prior Knowledge in Language Models},
  booktitle    = {Proceedings of the 62nd Annual Meeting of the Association for Computational
                  Linguistics (Volume 1: Long Papers), {ACL} 2024, Bangkok, Thailand,
                  August 11-16, 2024},
  pages        = {13211--13235},
  publisher    = {Association for Computational Linguistics},
  year         = {2024},
  url          = {https://doi.org/10.18653/v1/2024.acl-long.714},
  doi          = {10.18653/V1/2024.ACL-LONG.714},
  bibsource    = {dblp computer science bibliography, https://dblp.org}
}

@inproceedings{DBLP:conf/iclr/0002WSLCNCZ23,
  author       = {Xuezhi Wang and
                  Jason Wei and
                  Dale Schuurmans and
                  Quoc V. Le and
                  Ed H. Chi and
                  Sharan Narang and
                  Aakanksha Chowdhery and
                  Denny Zhou},
  title        = {Self-Consistency Improves Chain of Thought Reasoning in Language Models},
  booktitle    = {The Eleventh International Conference on Learning Representations,
                  {ICLR} 2023, Kigali, Rwanda, May 1-5, 2023},
  publisher    = {OpenReview.net},
  year         = {2023},
  url          = {https://openreview.net/forum?id=1PL1NIMMrw},
  bibsource    = {dblp computer science bibliography, https://dblp.org}
}

@inproceedings{DBLP:conf/iclr/ZhouSHWS0SCBLC23,
  author       = {Denny Zhou and
                  Nathanael Sch{\"{a}}rli and
                  Le Hou and
                  Jason Wei and
                  Nathan Scales and
                  Xuezhi Wang and
                  Dale Schuurmans and
                  Claire Cui and
                  Olivier Bousquet and
                  Quoc V. Le and
                  Ed H. Chi},
  title        = {Least-to-Most Prompting Enables Complex Reasoning in Large Language
                  Models},
  booktitle    = {The Eleventh International Conference on Learning Representations,
                  {ICLR} 2023, Kigali, Rwanda, May 1-5, 2023},
  publisher    = {OpenReview.net},
  year         = {2023},
  url          = {https://openreview.net/forum?id=WZH7099tgfM},
  bibsource    = {dblp computer science bibliography, https://dblp.org}
}

@inproceedings{DBLP:conf/nips/Wei0SBIXCLZ22,
  author       = {Jason Wei and
                  Xuezhi Wang and
                  Dale Schuurmans and
                  Maarten Bosma and
                  Brian Ichter and
                  Fei Xia and
                  Ed H. Chi and
                  Quoc V. Le and
                  Denny Zhou},
  editor       = {Sanmi Koyejo and
                  S. Mohamed and
                  A. Agarwal and
                  Danielle Belgrave and
                  K. Cho and
                  A. Oh},
  title        = {Chain-of-Thought Prompting Elicits Reasoning in Large Language Models},
  booktitle    = {Advances in Neural Information Processing Systems 35: Annual Conference
                  on Neural Information Processing Systems 2022, NeurIPS 2022, New Orleans,
                  LA, USA, November 28 - December 9, 2022},
  year         = {2022},
  url          = {http://papers.nips.cc/paper\_files/paper/2022/hash/9d5609613524ecf4f15af0f7b31abca4-Abstract-Conference.html},
  bibsource    = {dblp computer science bibliography, https://dblp.org}
}

@inproceedings{DBLP:conf/nips/YeCDD23,
  author       = {Xi Ye and
                  Qiaochu Chen and
                  Isil Dillig and
                  Greg Durrett},
  editor       = {Alice Oh and
                  Tristan Naumann and
                  Amir Globerson and
                  Kate Saenko and
                  Moritz Hardt and
                  Sergey Levine},
  title        = {SatLM: Satisfiability-Aided Language Models Using Declarative Prompting},
  booktitle    = {Advances in Neural Information Processing Systems 36: Annual Conference
                  on Neural Information Processing Systems 2023, NeurIPS 2023, New Orleans,
                  LA, USA, December 10 - 16, 2023},
  year         = {2023},
  url          = {http://papers.nips.cc/paper\_files/paper/2023/hash/8e9c7d4a48bdac81a58f983a64aaf42b-Abstract-Conference.html},
  bibsource    = {dblp computer science bibliography, https://dblp.org}
}

@inproceedings{DBLP:conf/nips/SaparovPPJKK023,
  author       = {Abulhair Saparov and
                  Richard Yuanzhe Pang and
                  Vishakh Padmakumar and
                  Nitish Joshi and
                  Mehran Kazemi and
                  Najoung Kim and
                  He He},
  editor       = {Alice Oh and
                  Tristan Naumann and
                  Amir Globerson and
                  Kate Saenko and
                  Moritz Hardt and
                  Sergey Levine},
  title        = {Testing the General Deductive Reasoning Capacity of Large Language
                  Models Using {OOD} Examples},
  booktitle    = {Advances in Neural Information Processing Systems 36: Annual Conference
                  on Neural Information Processing Systems 2023, NeurIPS 2023, New Orleans,
                  LA, USA, December 10 - 16, 2023},
  year         = {2023},
  url          = {http://papers.nips.cc/paper\_files/paper/2023/hash/09425891e393e64b0535194a81ba15b7-Abstract-Conference.html},
  bibsource    = {dblp computer science bibliography, https://dblp.org}
}

@inproceedings{DBLP:conf/emnlp/YangSWQZWRWL25,
  author       = {Yiran Yang and
                  Haifeng Sun and
                  Jingyu Wang and
                  Qi Qi and
                  Zirui Zhuang and
                  Huazheng Wang and
                  Pengfei Ren and
                  Jing Wang and
                  Jianxin Liao},
  editor       = {Christos Christodoulopoulos and
                  Tanmoy Chakraborty and
                  Carolyn Rose and
                  Violet Peng},
  title        = {Unveiling Internal Reasoning Modes in LLMs: {A} Deep Dive into Latent
                  Reasoning vs. Factual Shortcuts with Attribute Rate Ratio},
  booktitle    = {Proceedings of the 2025 Conference on Empirical Methods in Natural
                  Language Processing, {EMNLP} 2025, Suzhou, China, November 4-9, 2025},
  pages        = {2186--2206},
  publisher    = {Association for Computational Linguistics},
  year         = {2025},
  url          = {https://doi.org/10.18653/v1/2025.emnlp-main.111},
  doi          = {10.18653/V1/2025.EMNLP-MAIN.111},
  bibsource    = {dblp computer science bibliography, https://dblp.org}
}

@inproceedings{DBLP:conf/emnlp/YuanZZZL24,
  author       = {Yu Yuan and
                  Lili Zhao and
                  Kai Zhang and
                  Guangting Zheng and
                  Qi Liu},
  editor       = {Yaser Al{-}Onaizan and
                  Mohit Bansal and
                  Yun{-}Nung Chen},
  title        = {Do LLMs Overcome Shortcut Learning? An Evaluation of Shortcut Challenges
                  in Large Language Models},
  booktitle    = {Proceedings of the 2024 Conference on Empirical Methods in Natural
                  Language Processing, {EMNLP} 2024, Miami, FL, USA, November 12-16,
                  2024},
  pages        = {12188--12200},
  publisher    = {Association for Computational Linguistics},
  year         = {2024},
  url          = {https://doi.org/10.18653/v1/2024.emnlp-main.679},
  doi          = {10.18653/V1/2024.EMNLP-MAIN.679},
  bibsource    = {dblp computer science bibliography, https://dblp.org}
}

@inproceedings{DBLP:conf/naacl/ZhouZAKZ24,
  author       = {Yue Zhou and
                  Yada Zhu and
                  Diego Antognini and
                  Yoon Kim and
                  Yang Zhang},
  editor       = {Kevin Duh and
                  Helena G{\'{o}}mez{-}Adorno and
                  Steven Bethard},
  title        = {Paraphrase and Solve: Exploring and Exploiting the Impact of Surface
                  Form on Mathematical Reasoning in Large Language Models},
  booktitle    = {Proceedings of the 2024 Conference of the North American Chapter of
                  the Association for Computational Linguistics: Human Language Technologies
                  (Volume 1: Long Papers), {NAACL} 2024, Mexico City, Mexico, June 16-21,
                  2024},
  pages        = {2793--2804},
  publisher    = {Association for Computational Linguistics},
  year         = {2024},
  url          = {https://doi.org/10.18653/v1/2024.naacl-long.153},
  doi          = {10.18653/V1/2024.NAACL-LONG.153},
  bibsource    = {dblp computer science bibliography, https://dblp.org}
}

@inproceedings{DBLP:conf/iclr/MirzadehASTBF25,
  author       = {Iman Mirzadeh and
                  Keivan Alizadeh and
                  Hooman Shahrokhi and
                  Oncel Tuzel and
                  Samy Bengio and
                  Mehrdad Farajtabar},
  title        = {GSM-Symbolic: Understanding the Limitations of Mathematical Reasoning
                  in Large Language Models},
  booktitle    = {The Thirteenth International Conference on Learning Representations,
                  {ICLR} 2025, Singapore, April 24-28, 2025},
  publisher    = {OpenReview.net},
  year         = {2025},
  url          = {https://openreview.net/forum?id=AjXkRZIvjB},
  bibsource    = {dblp computer science bibliography, https://dblp.org}
}

@book{copi1990introduction,
  title={Introduction to Logic},
  author={Copi, I.M. and Cohen, C.},
  isbn={9780023250354},
  lccn={89037742},
  series={Maxwell Macmillan international editions},
  url={https://books.google.com/books?id=UnIbAQAAMAAJ},
  year={1990},
  publisher={Macmillan}
}

@book{10.5555/2481002,
author = {Genesereth, Michael and Kao, Eric},
title = {Introduction to Logic},
year = {2012},
isbn = {1627050051},
publisher = {Morgan \& Claypool Publishers}
}

@inproceedings{DBLP:conf/stoc/BabaiL83,
  author       = {L{\'{a}}szl{\'{o}} Babai and
                  Eugene M. Luks},
  editor       = {David S. Johnson and
                  Ronald Fagin and
                  Michael L. Fredman and
                  David Harel and
                  Richard M. Karp and
                  Nancy A. Lynch and
                  Christos H. Papadimitriou and
                  Ronald L. Rivest and
                  Walter L. Ruzzo and
                  Joel I. Seiferas},
  title        = {Canonical Labeling of Graphs},
  booktitle    = {Proceedings of the 15th Annual {ACM} Symposium on Theory of Computing,
                  25-27 April, 1983, Boston, Massachusetts, {USA}},
  pages        = {171--183},
  publisher    = {{ACM}},
  year         = {1983},
  url          = {https://doi.org/10.1145/800061.808746},
  doi          = {10.1145/800061.808746},
  bibsource    = {dblp computer science bibliography, https://dblp.org}
}

@inproceedings{10.1145/123186.123222,
author = {Brace, Karl S. and Rudell, Richard L. and Bryant, Randal E.},
title = {Efficient implementation of a BDD package},
year = {1991},
isbn = {0897913639},
publisher = {Association for Computing Machinery},
address = {New York, NY, USA},
url = {https://doi.org/10.1145/123186.123222},
doi = {10.1145/123186.123222},
booktitle = {Proceedings of the 27th ACM/IEEE Design Automation Conference},
pages = {40–45},
numpages = {6},
location = {Orlando, Florida, USA},
series = {DAC '90}
}

@inproceedings{DBLP:conf/emnlp/CaoLDYZZSLHG25,
  author       = {Chuxue Cao and
                  Mengze Li and
                  Juntao Dai and
                  Jinluan Yang and
                  Zijian Zhao and
                  Shengyu Zhang and
                  Weijie Shi and
                  Chengzhong Liu and
                  Sirui Han and
                  Yike Guo},
  editor       = {Christos Christodoulopoulos and
                  Tanmoy Chakraborty and
                  Carolyn Rose and
                  Violet Peng},
  title        = {Towards Advanced Mathematical Reasoning for LLMs via First-Order Logic
                  Theorem Proving},
  booktitle    = {Proceedings of the 2025 Conference on Empirical Methods in Natural
                  Language Processing, {EMNLP} 2025, Suzhou, China, November 4-9, 2025},
  pages        = {12429--12449},
  publisher    = {Association for Computational Linguistics},
  year         = {2025},
  url          = {https://doi.org/10.18653/v1/2025.emnlp-main.628},
  doi          = {10.18653/V1/2025.EMNLP-MAIN.628},
  bibsource    = {dblp computer science bibliography, https://dblp.org}
}

@inproceedings{DBLP:conf/nips/HeDDG24,
  author       = {Tianyu He and
                  Darshil Doshi and
                  Aritra Das and
                  Andrey Gromov},
  editor       = {Amir Globersons and
                  Lester Mackey and
                  Danielle Belgrave and
                  Angela Fan and
                  Ulrich Paquet and
                  Jakub M. Tomczak and
                  Cheng Zhang},
  title        = {Learning to grok: Emergence of in-context learning and skill composition
                  in modular arithmetic tasks},
  booktitle    = {Advances in Neural Information Processing Systems 38: Annual Conference
                  on Neural Information Processing Systems 2024, NeurIPS 2024, Vancouver,
                  BC, Canada, December 10 - 15, 2024},
  year         = {2024},
  url          = {http://papers.nips.cc/paper\_files/paper/2024/hash/17d60fef592086d1a5cb136f1946df59-Abstract-Conference.html},
  bibsource    = {dblp computer science bibliography, https://dblp.org}
}

@inproceedings{DBLP:conf/iclr/ParkLLYONWT25,
  author       = {Core Francisco Park and
                  Andrew Lee and
                  Ekdeep Singh Lubana and
                  Yongyi Yang and
                  Maya Okawa and
                  Kento Nishi and
                  Martin Wattenberg and
                  Hidenori Tanaka},
  title        = {{ICLR:} In-Context Learning of Representations},
  booktitle    = {The Thirteenth International Conference on Learning Representations,
                  {ICLR} 2025, Singapore, April 24-28, 2025},
  publisher    = {OpenReview.net},
  year         = {2025},
  url          = {https://openreview.net/forum?id=pXlmOmlHJZ},
  bibsource    = {dblp computer science bibliography, https://dblp.org}
}

@inproceedings{DBLP:conf/iccv/TragerPZABS23,
  author       = {Matthew Trager and
                  Pramuditha Perera and
                  Luca Zancato and
                  Alessandro Achille and
                  Parminder Bhatia and
                  Stefano Soatto},
  title        = {Linear Spaces of Meanings: Compositional Structures in Vision-Language
                  Models},
  booktitle    = {{IEEE/CVF} International Conference on Computer Vision, {ICCV} 2023,
                  Paris, France, October 1-6, 2023},
  pages        = {15349--15358},
  publisher    = {{IEEE}},
  year         = {2023},
  url          = {https://doi.org/10.1109/ICCV51070.2023.01412},
  doi          = {10.1109/ICCV51070.2023.01412},
  bibsource    = {dblp computer science bibliography, https://dblp.org}
}

@inproceedings{DBLP:conf/nips/YanSKRH20,
  author       = {Yujun Yan and
                  Kevin Swersky and
                  Danai Koutra and
                  Parthasarathy Ranganathan and
                  Milad Hashemi},
  editor       = {Hugo Larochelle and
                  Marc'Aurelio Ranzato and
                  Raia Hadsell and
                  Maria{-}Florina Balcan and
                  Hsuan{-}Tien Lin},
  title        = {Neural Execution Engines: Learning to Execute Subroutines},
  booktitle    = {Advances in Neural Information Processing Systems 33: Annual Conference
                  on Neural Information Processing Systems 2020, NeurIPS 2020, December
                  6-12, 2020, virtual},
  year         = {2020},
  url          = {https://proceedings.neurips.cc/paper/2020/hash/c8b9abffb45bf79a630fb613dcd23449-Abstract.html},
  bibsource    = {dblp computer science bibliography, https://dblp.org}
}

@inproceedings{DBLP:conf/icml/ParkCV24,
  author       = {Kiho Park and
                  Yo Joong Choe and
                  Victor Veitch},
  editor       = {Ruslan Salakhutdinov and
                  Zico Kolter and
                  Katherine A. Heller and
                  Adrian Weller and
                  Nuria Oliver and
                  Jonathan Scarlett and
                  Felix Berkenkamp},
  title        = {The Linear Representation Hypothesis and the Geometry of Large Language
                  Models},
  booktitle    = {Forty-first International Conference on Machine Learning, {ICML} 2024,
                  Vienna, Austria, July 21-27, 2024},
  series       = {Proceedings of Machine Learning Research},
  pages        = {39643--39666},
  publisher    = {{PMLR} / OpenReview.net},
  year         = {2024},
  url          = {https://proceedings.mlr.press/v235/park24c.html},
  bibsource    = {dblp computer science bibliography, https://dblp.org}
}

@inproceedings{DBLP:conf/nips/AlayracDLMBHLMM22,
  author       = {Jean{-}Baptiste Alayrac and
                  Jeff Donahue and
                  Pauline Luc and
                  Antoine Miech and
                  Iain Barr and
                  Yana Hasson and
                  Karel Lenc and
                  Arthur Mensch and
                  Katherine Millican and
                  Malcolm Reynolds and
                  Roman Ring and
                  Eliza Rutherford and
                  Serkan Cabi and
                  Tengda Han and
                  Zhitao Gong and
                  Sina Samangooei and
                  Marianne Monteiro and
                  Jacob L. Menick and
                  Sebastian Borgeaud and
                  Andy Brock and
                  Aida Nematzadeh and
                  Sahand Sharifzadeh and
                  Mikolaj Binkowski and
                  Ricardo Barreira and
                  Oriol Vinyals and
                  Andrew Zisserman and
                  Kar{\'{e}}n Simonyan},
  editor       = {Sanmi Koyejo and
                  S. Mohamed and
                  A. Agarwal and
                  Danielle Belgrave and
                  K. Cho and
                  A. Oh},
  title        = {Flamingo: a Visual Language Model for Few-Shot Learning},
  booktitle    = {Advances in Neural Information Processing Systems 35: Annual Conference
                  on Neural Information Processing Systems 2022, NeurIPS 2022, New Orleans,
                  LA, USA, November 28 - December 9, 2022},
  year         = {2022},
  url          = {http://papers.nips.cc/paper\_files/paper/2022/hash/960a172bc7fbf0177ccccbb411a7d800-Abstract-Conference.html},
  bibsource    = {dblp computer science bibliography, https://dblp.org}
}

@article{DBLP:journals/corr/abs-1807-03748,
  author       = {A{\"{a}}ron van den Oord and
                  Yazhe Li and
                  Oriol Vinyals},
  title        = {Representation Learning with Contrastive Predictive Coding},
  journal      = {CoRR},
  volume       = {abs/1807.03748},
  year         = {2018},
  url          = {http://arxiv.org/abs/1807.03748},
  eprinttype   = {arXiv},
  eprint       = {1807.03748},
  bibsource    = {dblp computer science bibliography, https://dblp.org}
}

@inproceedings{DBLP:conf/iclr/RobinsonCSJ21,
  author       = {Joshua David Robinson and
                  Ching{-}Yao Chuang and
                  Suvrit Sra and
                  Stefanie Jegelka},
  title        = {Contrastive Learning with Hard Negative Samples},
  booktitle    = {9th International Conference on Learning Representations, {ICLR} 2021,
                  Virtual Event, Austria, May 3-7, 2021},
  publisher    = {OpenReview.net},
  year         = {2021},
  url          = {https://openreview.net/forum?id=CR1XOQ0UTh-},
  bibsource    = {dblp computer science bibliography, https://dblp.org}
}

@inproceedings{DBLP:conf/icml/0001I20,
  author       = {Tongzhou Wang and
                  Phillip Isola},
  title        = {Understanding Contrastive Representation Learning through Alignment
                  and Uniformity on the Hypersphere},
  booktitle    = {Proceedings of the 37th International Conference on Machine Learning,
                  {ICML} 2020, 13-18 July 2020, Virtual Event},
  series       = {Proceedings of Machine Learning Research},
  pages        = {9929--9939},
  publisher    = {{PMLR}},
  year         = {2020},
  url          = {http://proceedings.mlr.press/v119/wang20k.html},
  bibsource    = {dblp computer science bibliography, https://dblp.org}
}

@inproceedings{DBLP:conf/emnlp/HanS0QRZCPQBSWS24,
  author       = {Simeng Han and
                  Hailey Schoelkopf and
                  Yilun Zhao and
                  Zhenting Qi and
                  Martin Riddell and
                  Wenfei Zhou and
                  James Coady and
                  David Peng and
                  Yujie Qiao and
                  Luke Benson and
                  Lucy Sun and
                  Alexander Wardle{-}Solano and
                  Hannah Szab{\'{o}} and
                  Ekaterina Zubova and
                  Matthew Burtell and
                  Jonathan Fan and
                  Yixin Liu and
                  Brian Wong and
                  Malcolm Sailor and
                  Ansong Ni and
                  Linyong Nan and
                  Jungo Kasai and
                  Tao Yu and
                  Rui Zhang and
                  Alexander R. Fabbri and
                  Wojciech Kryscinski and
                  Semih Yavuz and
                  Ye Liu and
                  Xi Victoria Lin and
                  Shafiq Joty and
                  Yingbo Zhou and
                  Caiming Xiong and
                  Rex Ying and
                  Arman Cohan and
                  Dragomir Radev},
  editor       = {Yaser Al{-}Onaizan and
                  Mohit Bansal and
                  Yun{-}Nung Chen},
  title        = {{FOLIO:} Natural Language Reasoning with First-Order Logic},
  booktitle    = {Proceedings of the 2024 Conference on Empirical Methods in Natural
                  Language Processing, {EMNLP} 2024, Miami, FL, USA, November 12-16,
                  2024},
  pages        = {22017--22031},
  publisher    = {Association for Computational Linguistics},
  year         = {2024},
  url          = {https://doi.org/10.18653/v1/2024.emnlp-main.1229},
  doi          = {10.18653/V1/2024.EMNLP-MAIN.1229},
  bibsource    = {dblp computer science bibliography, https://dblp.org}
}

@article{DBLP:journals/corr/SchulmanWDRK17,
  author       = {John Schulman and
                  Filip Wolski and
                  Prafulla Dhariwal and
                  Alec Radford and
                  Oleg Klimov},
  title        = {Proximal Policy Optimization Algorithms},
  journal      = {CoRR},
  volume       = {abs/1707.06347},
  year         = {2017},
  url          = {http://arxiv.org/abs/1707.06347},
  eprinttype   = {arXiv},
  eprint       = {1707.06347},
  bibsource    = {dblp computer science bibliography, https://dblp.org}
}

@article{DBLP:journals/corr/abs-2407-21783,
  author       = {Llama Team},
  title        = {The Llama 3 Herd of Models},
  journal      = {CoRR},
  volume       = {abs/2407.21783},
  year         = {2024},
  url          = {https://doi.org/10.48550/arXiv.2407.21783},
  doi          = {10.48550/ARXIV.2407.21783},
  eprinttype   = {arXiv},
  eprint       = {2407.21783},
  bibsource    = {dblp computer science bibliography, https://dblp.org}
}

@misc{openai2025o3o4mini,
  author       = {{OpenAI}},
  title        = {Introducing OpenAI o3 and o4-mini},
  year         = {2025},
  month        = apr,
  howpublished = {\url{https://openai.com/index/introducing-o3-and-o4-mini/}},
}

@article{DBLP:journals/csur/Bryant92,
  author       = {Randal E. Bryant},
  title        = {Symbolic Boolean Manipulation with Ordered Binary-Decision Diagrams},
  journal      = {{ACM} Comput. Surv.},
  volume       = {24},
  number       = {3},
  pages        = {293--318},
  year         = {1992},
  url          = {https://doi.org/10.1145/136035.136043},
  doi          = {10.1145/136035.136043},
  bibsource    = {dblp computer science bibliography, https://dblp.org}
}

@article{DBLP:journals/tc/Bryant86,
  author       = {Randal E. Bryant},
  title        = {Graph-Based Algorithms for Boolean Function Manipulation},
  journal      = {{IEEE} Trans. Computers},
  volume       = {35},
  number       = {8},
  pages        = {677--691},
  year         = {1986},
  url          = {https://doi.org/10.1109/TC.1986.1676819},
  doi          = {10.1109/TC.1986.1676819},
  bibsource    = {dblp computer science bibliography, https://dblp.org}
}

@article{DBLP:journals/corr/abs-2505-09388,
  author       = {Qwen Team},
  title        = {Qwen3 Technical Report},
  journal      = {CoRR},
  volume       = {abs/2505.09388},
  year         = {2025},
  url          = {https://doi.org/10.48550/arXiv.2505.09388},
  doi          = {10.48550/ARXIV.2505.09388},
  eprinttype   = {arXiv},
  eprint       = {2505.09388},
  bibsource    = {dblp computer science bibliography, https://dblp.org}
}

@article{DBLP:journals/corr/abs-2507-06261,
  author       = {Gemini Team},
  title        = {Gemini 2.5: Pushing the Frontier with Advanced Reasoning, Multimodality,
                  Long Context, and Next Generation Agentic Capabilities},
  journal      = {CoRR},
  volume       = {abs/2507.06261},
  year         = {2025},
  url          = {https://doi.org/10.48550/arXiv.2507.06261},
  doi          = {10.48550/ARXIV.2507.06261},
  eprinttype   = {arXiv},
  eprint       = {2507.06261},
  bibsource    = {dblp computer science bibliography, https://dblp.org}
}

@article{DBLP:journals/logcom/Fitting02,
  author       = {Melvin Fitting},
  title        = {Modal Logics Between Propositional and First-order},
  journal      = {J. Log. Comput.},
  volume       = {12},
  number       = {6},
  pages        = {1017--1026},
  year         = {2002},
  url          = {https://doi.org/10.1093/logcom/12.6.1017},
  doi          = {10.1093/LOGCOM/12.6.1017},
  bibsource    = {dblp computer science bibliography, https://dblp.org}
}

@book{DBLP:books/daglib/0082098,
  author       = {Melvin Fitting},
  title        = {First-Order Logic and Automated Theorem Proving, Second Edition},
  series       = {Graduate Texts in Computer Science},
  publisher    = {Springer},
  year         = {1996},
  url          = {https://doi.org/10.1007/978-1-4612-2360-3},
  doi          = {10.1007/978-1-4612-2360-3},
  isbn         = {978-1-4612-7515-2},
  bibsource    = {dblp computer science bibliography, https://dblp.org}
}

@inproceedings{DBLP:conf/icml/JiaYXCPPLSLD21,
  author       = {Chao Jia and
                  Yinfei Yang and
                  Ye Xia and
                  Yi{-}Ting Chen and
                  Zarana Parekh and
                  Hieu Pham and
                  Quoc V. Le and
                  Yun{-}Hsuan Sung and
                  Zhen Li and
                  Tom Duerig},
  editor       = {Marina Meila and
                  Tong Zhang},
  title        = {Scaling Up Visual and Vision-Language Representation Learning With
                  Noisy Text Supervision},
  booktitle    = {Proceedings of the 38th International Conference on Machine Learning,
                  {ICML} 2021, 18-24 July 2021, Virtual Event},
  series       = {Proceedings of Machine Learning Research},
  pages        = {4904--4916},
  publisher    = {{PMLR}},
  year         = {2021},
  url          = {http://proceedings.mlr.press/v139/jia21b.html},
  bibsource    = {dblp computer science bibliography, https://dblp.org}
}

@inproceedings{
li2026table,
title={Table as a Modality for Large Language Models},
author={Liyao Li and Chao Ye and Wentao Ye and Yifei Sun and Zhe Jiang and Haobo Wang and Jiaming Tian and Yiming Zhang and NINGTAO WANG and Xing Fu and Gang Chen and Junbo Zhao},
booktitle={The Thirty-ninth Annual Conference on Neural Information Processing Systems},
year={2025},
url={https://openreview.net/forum?id=kurEZdWU9G}
}

@inproceedings{DBLP:conf/iclr/BerglundTKBSKE24,
  author       = {Lukas Berglund and
                  Meg Tong and
                  Maximilian Kaufmann and
                  Mikita Balesni and
                  Asa Cooper Stickland and
                  Tomasz Korbak and
                  Owain Evans},
  title        = {The Reversal Curse: LLMs trained on "A is B" fail to learn "B is A"},
  booktitle    = {The Twelfth International Conference on Learning Representations,
                  {ICLR} 2024, Vienna, Austria, May 7-11, 2024},
  publisher    = {OpenReview.net},
  year         = {2024},
  url          = {https://openreview.net/forum?id=GPKTIktA0k},
  bibsource    = {dblp computer science bibliography, https://dblp.org}
}

@inproceedings{
pei2024exploiting,
title={Exploiting Code Symmetries for Learning Program Semantics},
author={Kexin Pei and Weichen Li and Qirui Jin and Shuyang Liu and Scott Geng and Lorenzo Cavallaro and Junfeng Yang and Suman Jana},
booktitle={Forty-first International Conference on Machine Learning},
year={2024},
url={https://openreview.net/forum?id=OLvgrLtv6J}
}

@article{DBLP:journals/corr/abs-2509-25851,
  author       = {Jundong Xu and
                  Hao Fei and
                  Yuhui Zhang and
                  Liangming Pan and
                  Qijun Huang and
                  Qian Liu and
                  Preslav Nakov and
                  Min{-}Yen Kan and
                  William Yang Wang and
                  Mong{-}Li Lee and
                  Wynne Hsu},
  title        = {MuSLR: Multimodal Symbolic Logical Reasoning},
  journal      = {CoRR},
  volume       = {abs/2509.25851},
  year         = {2025},
  url          = {https://doi.org/10.48550/arXiv.2509.25851},
  doi          = {10.48550/ARXIV.2509.25851},
  eprinttype   = {arXiv},
  eprint       = {2509.25851},
  bibsource    = {dblp computer science bibliography, https://dblp.org}
}

@article{DBLP:journals/corr/abs-2412-15115,
  author       = {An Yang and
                  Baosong Yang and
                  Beichen Zhang and
                  Binyuan Hui and
                  Bo Zheng and
                  Bowen Yu and
                  Chengyuan Li and
                  Dayiheng Liu and
                  Fei Huang and
                  Haoran Wei and
                  Huan Lin and
                  Jian Yang and
                  Jianhong Tu and
                  Jianwei Zhang and
                  Jianxin Yang and
                  Jiaxi Yang and
                  Jingren Zhou and
                  Junyang Lin and
                  Kai Dang and
                  Keming Lu and
                  Keqin Bao and
                  Kexin Yang and
                  Le Yu and
                  Mei Li and
                  Mingfeng Xue and
                  Pei Zhang and
                  Qin Zhu and
                  Rui Men and
                  Runji Lin and
                  Tianhao Li and
                  Tingyu Xia and
                  Xingzhang Ren and
                  Xuancheng Ren and
                  Yang Fan and
                  Yang Su and
                  Yichang Zhang and
                  Yu Wan and
                  Yuqiong Liu and
                  Zeyu Cui and
                  Zhenru Zhang and
                  Zihan Qiu},
  title        = {Qwen2.5 Technical Report},
  journal      = {CoRR},
  volume       = {abs/2412.15115},
  year         = {2024},
  url          = {https://doi.org/10.48550/arXiv.2412.15115},
  doi          = {10.48550/ARXIV.2412.15115},
  eprinttype   = {arXiv},
  eprint       = {2412.15115},
  bibsource    = {dblp computer science bibliography, https://dblp.org}
}

@inproceedings{Lin2024GT2VecLL,
  title={GT2Vec: Large Language Models as Multi-Modal Encoders for Text and Graph-Structured Data},
  author={Jiacheng Lin and Kun Qian and Haoyu Han and Nurendra Choudhary and Tianxin Wei and Zhongruo Wang and Sahika Genc and E-Wen Huang and Sheng Wang and Karthik Subbian and Danai Koutra and Jimeng Sun},
  year={2024},
  url={https://api.semanticscholar.org/CorpusID:273350888}
}

@inproceedings{DBLP:conf/iclr/0003AEAAZW25,
  author       = {Xinyi Wang and
                  Antonis Antoniades and
                  Yanai Elazar and
                  Alfonso Amayuelas and
                  Alon Albalak and
                  Kexun Zhang and
                  William Yang Wang},
  title        = {Generalization v.s. Memorization: Tracing Language Models' Capabilities
                  Back to Pretraining Data},
  booktitle    = {The Thirteenth International Conference on Learning Representations,
                  {ICLR} 2025, Singapore, April 24-28, 2025},
  publisher    = {OpenReview.net},
  year         = {2025},
  url          = {https://openreview.net/forum?id=IQxBDLmVpT},
  bibsource    = {dblp computer science bibliography, https://dblp.org}
}

@inproceedings{DBLP:conf/nips/LiuLWL23a,
  author       = {Haotian Liu and
                  Chunyuan Li and
                  Qingyang Wu and
                  Yong Jae Lee},
  editor       = {Alice Oh and
                  Tristan Naumann and
                  Amir Globerson and
                  Kate Saenko and
                  Moritz Hardt and
                  Sergey Levine},
  title        = {Visual Instruction Tuning},
  booktitle    = {Advances in Neural Information Processing Systems 36: Annual Conference
                  on Neural Information Processing Systems 2023, NeurIPS 2023, New Orleans,
                  LA, USA, December 10 - 16, 2023},
  year         = {2023},
  url          = {http://papers.nips.cc/paper\_files/paper/2023/hash/6dcf277ea32ce3288914faf369fe6de0-Abstract-Conference.html},
  bibsource    = {dblp computer science bibliography, https://dblp.org}
}

@inproceedings{DBLP:conf/iclr/Chen0CPPSGGMB0P23,
  author       = {Xi Chen and
                  Xiao Wang and
                  Soravit Changpinyo and
                  A. J. Piergiovanni and
                  Piotr Padlewski and
                  Daniel Salz and
                  Sebastian Goodman and
                  Adam Grycner and
                  Basil Mustafa and
                  Lucas Beyer and
                  Alexander Kolesnikov and
                  Joan Puigcerver and
                  Nan Ding and
                  Keran Rong and
                  Hassan Akbari and
                  Gaurav Mishra and
                  Linting Xue and
                  Ashish V. Thapliyal and
                  James Bradbury and
                  Weicheng Kuo},
  title        = {PaLI: {A} Jointly-Scaled Multilingual Language-Image Model},
  booktitle    = {The Eleventh International Conference on Learning Representations,
                  {ICLR} 2023, Kigali, Rwanda, May 1-5, 2023},
  publisher    = {OpenReview.net},
  year         = {2023},
  url          = {https://openreview.net/forum?id=mWVoBz4W0u},
  bibsource    = {dblp computer science bibliography, https://dblp.org}
}

@inproceedings{DBLP:conf/icml/RadfordKHRGASAM21,
  author       = {Alec Radford and
                  Jong Wook Kim and
                  Chris Hallacy and
                  Aditya Ramesh and
                  Gabriel Goh and
                  Sandhini Agarwal and
                  Girish Sastry and
                  Amanda Askell and
                  Pamela Mishkin and
                  Jack Clark and
                  Gretchen Krueger and
                  Ilya Sutskever},
  editor       = {Marina Meila and
                  Tong Zhang},
  title        = {Learning Transferable Visual Models From Natural Language Supervision},
  booktitle    = {Proceedings of the 38th International Conference on Machine Learning,
                  {ICML} 2021, 18-24 July 2021, Virtual Event},
  series       = {Proceedings of Machine Learning Research},
  pages        = {8748--8763},
  publisher    = {{PMLR}},
  year         = {2021},
  url          = {http://proceedings.mlr.press/v139/radford21a.html},
  bibsource    = {dblp computer science bibliography, https://dblp.org}
}

@inproceedings{10.5555/1792734.1792766,
author = {De Moura, Leonardo and Bj\o{}rner, Nikolaj},
title = {Z3: an efficient SMT solver},
year = {2008},
isbn = {3540787992},
publisher = {Springer-Verlag},
address = {Berlin, Heidelberg},
booktitle = {Proceedings of the Theory and Practice of Software, 14th International Conference on Tools and Algorithms for the Construction and Analysis of Systems},
pages = {337–340},
numpages = {4},
location = {Budapest, Hungary},
series = {TACAS'08/ETAPS'08}
}

@article{DeMoura2015Lean,
  title={The Lean Theorem Prover (System Description)},
  author={Leonardo De Moura and Soonho Kong and Jeremy Avigad and Floris van Doorn and Jakob von Raumer},
  journal={Automated Deduction – CADE-25},
  year={2015},
  volume={9195},
  pages={378--388},
  publisher={Springer}
}

@article{DBLP:journals/corr/abs-2402-03300,
  author       = {Zhihong Shao and
                  Peiyi Wang and
                  Qihao Zhu and
                  Runxin Xu and
                  Junxiao Song and
                  Mingchuan Zhang and
                  Y. K. Li and
                  Y. Wu and
                  Daya Guo},
  title        = {DeepSeekMath: Pushing the Limits of Mathematical Reasoning in Open
                  Language Models},
  journal      = {CoRR},
  volume       = {abs/2402.03300},
  year         = {2024},
  url          = {https://doi.org/10.48550/arXiv.2402.03300},
  doi          = {10.48550/ARXIV.2402.03300},
  eprinttype   = {arXiv},
  eprint       = {2402.03300},
  bibsource    = {dblp computer science bibliography, https://dblp.org}
}

@inproceedings{
olausson2023linc,
title={{LINC}: A Neurosymbolic Approach for Logical Reasoning by Combining Language Models with First-Order Logic Provers},
author={Theo X. Olausson and Alex Gu and Ben Lipkin and Cedegao E. Zhang and Armando Solar-Lezama and Joshua B. Tenenbaum and Roger P. Levy},
booktitle={The 2023 Conference on Empirical Methods in Natural Language Processing},
year={2023},
url={https://openreview.net/forum?id=h00GHjWDEp}
}

@inproceedings{pan-etal-2023-logic,
    title = "Logic-{LM}: Empowering Large Language Models with Symbolic Solvers for Faithful Logical Reasoning",
    author = "Pan, Liangming  and
      Albalak, Alon  and
      Wang, Xinyi  and
      Wang, William",
    editor = "Bouamor, Houda  and
      Pino, Juan  and
      Bali, Kalika",
    booktitle = "Findings of the Association for Computational Linguistics: EMNLP 2023",
    month = dec,
    year = "2023",
    address = "Singapore",
    publisher = "Association for Computational Linguistics",
    url = "https://aclanthology.org/2023.findings-emnlp.248/",
    doi = "10.18653/v1/2023.findings-emnlp.248",
    pages = "3806--3824"
}

@misc{McCune2005Prover9,
  title={Prover9 and Mace4},
  author={William McCune},
  year={2005},
  howpublished={\url{https://www.cs.unm.edu/~mccune/prover9/}}
}

@article{yan2020neural,
  title={Neural execution engines: Learning to execute subroutines},
  author={Yan, Yujun and Swersky, Kevin and Koutra, Danai and Ranganathan, Parthasarathy and Hashemi, Milad},
  journal={Advances in Neural Information Processing Systems},
  volume={33},
  pages={17298--17308},
  year={2020}
}

@misc{zeng2026sagemitigatinglonghorizonreasoning,
      title={SAGE: Mitigating Long-Horizon Reasoning Biases via Topological Guidance}, 
      author={Xinyue Zeng and Jiawei Zhang and Yujun Yan and Dawei Zhou},
      year={2026},
      eprint={2609.30192},
      archivePrefix={arXiv},
      primaryClass={cs.AI},
      url={https://arxiv.org/abs/2609.30192}, 
}

\newpage
\appendix

\section{Appendix}

\subsection{Logical Encoder Algorithm}
\label{appendix:algorithm}

\begin{algorithm*}[!h]
\caption{Algorithm for ROBDD Construction}
\label{alg:robdd_construction}
\begin{algorithmic}[1]
\Require Formal premises $\Gamma^{\mathrm{FL}}=\mathcal{F}\cup\mathcal{R}$,
entities $\mathcal{E}$, and ground query $\mathcal{Q}^{\mathrm{FL}}$
\Ensure ROBDD $G$ with root $\rho$, terminals $\tau_0,\tau_1$,
and low/high edge sets $E_0/E_1$

\Statex \textit{// Stage 1: Grounding}
\State $\Gamma_{\mathrm{g}}\leftarrow
\operatorname{Ground}_{\mathcal{E}}(\Gamma^{\mathrm{FL}})$
\Comment{ground all rules over the instance's entities}
\State $F\leftarrow\bigwedge_{g\in\Gamma_{\mathrm{g}}}g$
\Comment{each ground atom is a Boolean variable}

\Statex \textit{// Stage 2: Variable Ordering}
\State $L\leftarrow$
\Call{BackwardDFS}{$\operatorname{atoms}(\mathcal{Q}^{\mathrm{FL}}),\Gamma_{\mathrm{g}}$}
\Comment{recursively follow fact/rule dependencies; visit each atom once}
\State $(x_1,\ldots,x_m)\leftarrow
L\,\|\,\operatorname{Sort}\!\left(
\operatorname{atoms}(\Gamma_{\mathrm{g}})\setminus\operatorname{set}(L)
\right)$ \Comment{Use shared dependencies and fixed traversal/sorting orders across equivalent variants.}

\Statex \textit{// Stage 3: Shannon Expansion and Reduction}
\State Initialize $G$ with $\tau_0,\tau_1$ and $E_0=E_1=\varnothing$
\Function{Build}{$F,i$}
    \State \textbf{if} $F\equiv c\in\{0,1\}$
    \textbf{then return} $\tau_c$
    \State $u_b\leftarrow$
    \Call{Build}{$F|_{x_i=b},i+1$} for $b\in\{0,1\}$
    \Comment{Shannon expansion}
    \State \textbf{if} $u_0=u_1$ \textbf{then return} $u_0$
    \Comment{remove redundant nodes}
    \State \Return existing node $(x_i,u_0,u_1)$ in $G$,
    or create one with low/high successors $u_0/u_1$
    \Comment{merge isomorphic subgraphs}
\EndFunction
\State $\rho\leftarrow$ \Call{Build}{$F,1$}
\State \Return $G$ rooted at $\rho$
\end{algorithmic}
\end{algorithm*}

\begin{algorithm*}[!h]
\caption{Algorithm for Logical Encoder}
\label{alg:robdd_encoder}
\begin{algorithmic}[1]
\Require ROBDD graph $G$ with terminal nodes $\{\tau_0,\tau_1\}\subseteq V$; low/high branch sets $E_0/E_1$; target LLM embedding layer $\mathrm{Emb}(\cdot)$; edge-type-specific projections $\{\mathbf{W}_g^k,\mathbf{W}_m^k\}_{k\in\{0,1\}}$, where $\mathbf{W}_g^k$ produces the gate and $\mathbf{W}_m^k$ produces the message for edge type $k$

\Ensure Graph-level representation $\mathbf{h}_{G} \in \mathbb{R}^{d_g}$.

\Statex \textit{// Stage 1: Initialization}
\For{each $v \in V$}
    \State $\mathbf{h}_v \leftarrow \mathrm{Emb}(v)$
\EndFor

\Statex \textit{// Stage 2: Topological Ordering}
\State Compute in-degree for each node and sort $V$ into topological layers $L_1, L_2, \ldots, L_T$

\Statex \textit{// Stage 3: Layer-wise Message Passing}
\For{$t = 1$ to $T$}
    \For{each $v \in L_t$ with $\mathrm{in\text{-}degree}(v) = 0$ in remaining graph}
        \For{$k \in \{0, 1\}$}
            \State $m_k(v) \leftarrow \sum_{(u,v) \in E_k} \sigma(\mathbf{W}_g^k \mathbf{h}_u) \odot (\mathbf{W}_m^k \mathbf{h}_u)$
        \EndFor
        \State $\mathbf{h}_v \leftarrow \mathrm{GRU}\!\left(m_0(v) + m_1(v),\; \mathbf{h}_v\right)$
    \EndFor
    \State Remove processed nodes and update in-degrees
\EndFor

\Statex \textit{// Stage 4: Graph-level Readout}
\State $\mathbf{h}_{G}  \leftarrow \mathrm{MLP}\!\left([\mathbf{h}_{\tau_0} \| \mathbf{h}_{\tau_1}]\right)$

\State \Return $\mathbf{h}_{G} $
\end{algorithmic}
\end{algorithm*}

\subsection{Study of Efficiency}
\label{appendix:efficiency}

We next quantify the parameter overhead introduced by LoGIST on top of the base LLM. 
Let \(L_{\mathrm{LLM}}\) denote the number of decoder layers in the base LLM and \(d_t\) its hidden dimension. 
For logic fusion, we insert a gated cross-attention module every \(g\) decoder layers, resulting in 
\(\lceil L_{\mathrm{LLM}}/g \rceil\) cross-attention modules in total. 
For the logic encoder, let \(L_g\) denote the number of message-passing layers and \(d_g\) the graph hidden dimension, 
which is used for both node states \(\mathbf{h}_v\) and the graph-level readout \(\mathbf{h}_{{G}_i}\).

LoGIST introduces learnable parameters through three components: (i) the cross-attention modules, which incorporate the logic representation into the LLM hidden states; (ii) the graph encoder backbone, which maps the input logic graph into a structure-aware representation; and (iii) the graph-to-text aligner, implemented as a two-layer MLP, which projects graph representations into the LLM hidden space.

Each cross-attention module operates in dimension $d_t $ and contributes $4\,d_t^2 + 6\,d_t $ parameters. Summing over the $\lceil L_{\mathrm{LLM}}/g \rceil $ inserted modules gives
\[
P_{\mathrm{cross}}
= \left\lceil \frac{L_{\mathrm{LLM}}}{g} \right\rceil
\left(4\,d_{t}^2 + 6\,d_{t}\right).
\]
Each graph encoder layer contains two edge-type-specific linear transformations and one GRU-based update, contributing $8\,d_g^2 + 8\,d_g $ parameters. Summing over the $L_g $ layers gives
\[
P_{\mathrm{enc}}
= L_g\!\left(8\,d_g^2 + 8\,d_g\right).
\]
The aligner adds
\[
P_{\mathrm{align}}
= d_g\,d_{t} + d_{t}^2 + 2\,d_{t}.
\]
The total parameter overhead is therefore

\begin{align*}
P_{\textsc{LoGIST}}
&= P_{\mathrm{cross}} + P_{\mathrm{enc}} + P_{\mathrm{align}} \notag \\
&= 2\left\lceil \frac{L_{\mathrm{LLM}}}{g} \right\rceil
d_{t}(2\,d_{t} + 3)
+ 8\,L_g\,d_g(d_g + 1)
+ d_{t}(d_g + d_{t} + 2).
\end{align*}

Concretely, in our setup, \texttt{Llama-3.1-8B-Ins} introduces 312M additional parameters and \texttt{Qwen2.5-3B-Ins} introduces 101M, both amounting to only about 3\% of the full model size. Most of the added parameters are concentrated in the cross-attention modules rather than the graph encoder, so the design remains parameter-efficient. Moreover, since the training pipeline freezes the graph encoder by default, the number of trainable additional parameters is strictly smaller than the total overhead reported above.

We further compare the inference speed of each base model with its LoGIST-enhanced variant. Under identical generation configurations and on the same device, we report the mean generation throughput, measured in generated tokens per second (token/s), over 500 examples in Table~\ref{tab:inference_speed}.

\begin{table}[htbp]
\caption{{Inference-speed comparison between the original models and their LoGIST-enhanced variants. LoGIST introduces minimal additional overhead compared with the base model.}}
\vspace{10pt}
\centering
\renewcommand{\arraystretch}{1.25} % increase row height
\begin{tabular*}{0.65\linewidth}{@{\extracolsep{\fill}}lcc}
\toprule
\textbf{Model} & \texttt{Llama-3.1-8B-Ins} & \texttt{Qwen2.5-3B-Ins} \\
\midrule
\textbf{w/o LoGIST} & 58.20  (token/s) &  56.76 (token/s) \\
\textbf{w/ LoGIST}  & 56.99  (token/s) & 55.70 (token/s) \\
\bottomrule
\end{tabular*}
\label{tab:inference_speed}
\end{table}

This result is expected: the logical embedding remains static during generation and is computed only once at the prefilling stage, so the main runtime overhead comes from the cross-attention modules, which add less than 2\% extra cost. LoGIST is therefore an efficient framework for incorporating logical structure into the LLM to guide structure-aware reasoning.

\subsection{Training Setup}
\label{appendix:training_config}

LoGIST is algorithm-agnostic and can be plugged into existing LLM training pipelines. We apply it to \texttt{Llama-3.1-8B-Ins}~\citep{DBLP:journals/corr/abs-2407-21783} and \texttt{Qwen2.5-3B-Ins}~\citep{DBLP:journals/corr/abs-2412-15115} under three training paradigms: SFT with CoT supervision, PPO~\citep{DBLP:journals/corr/SchulmanWDRK17}, and GRPO~\citep{DBLP:journals/corr/abs-2402-03300}. For SFT, we use CoT responses as training labels; for PPO and GRPO, we use a binary reward based on exact matching of the final answer. Across all three paradigms, the logic encoder is kept frozen after its contrastive pretraining on the synthetic implicational dataset, and only the LLM and the logic fusion module are updated.

To isolate the contribution of LoGIST from dataset-specific tuning, we use the same hyperparameters across all datasets within each paradigm rather than tuning them per dataset. The full configurations for \texttt{Llama-3.1-8B-Ins} are reported in Table~\ref{tab:training_config}.

\begin{table}[htbp]
\centering
\small
\caption{\small{Training configurations for the three training paradigms (SFT, PPO, GRPO) on \texttt{Llama-3.1-8B-Ins}.}}
\label{tab:training_config}
\begin{subtable}[t]{0.32\textwidth}
\centering
\caption{SFT}
\label{tab:config_sft}
\begin{tabular}{lc}
\toprule
\textbf{Configuration} & \textbf{Value} \\
\midrule
Learning Rate (LR) & 1e-5 \\
Batch Size & 128 \\
Epochs & 3 \\
Initial Gate Value ($\alpha_\ell$) & 5e-2 \\
Layer Interval $g$ & 8 \\
Graph Dimension & 512 \\
Cross-Atten LR & 5e-6 \\
\bottomrule
\end{tabular}
\end{subtable}
\hfill
\begin{subtable}[t]{0.32\textwidth}
\centering
\caption{PPO}
\label{tab:config_ppo}
\begin{tabular}{lc}
\toprule
\textbf{Configuration} & \textbf{Value} \\
\midrule
Actor Learning Rate & 1e-6 \\
Critic Learning Rate & 5e-6 \\
Batch Size & 144 \\
KL Coefficient & 0.02 \\
Epochs & 3 \\
Initial Gate Value ($\alpha_\ell$) & 5e-2 \\
Layer Interval $g$ & 8 \\
Graph Dimension & 512 \\
Cross-Atten LR & 5e-6 \\
\bottomrule
\end{tabular}
\end{subtable}
\hfill
\begin{subtable}[t]{0.32\textwidth}
\centering
\caption{GRPO}
\label{tab:config_grpo}
\begin{tabular}{lc}
\toprule
\textbf{Configuration} & \textbf{Value} \\
\midrule
Actor Learning Rate & 1e-6 \\
Batch Size & 144 \\
Rollout Size & 5 \\
KL Coefficient & 0.02 \\
Epochs & 3 \\
Initial Gate Value ($\alpha_\ell$) & 5e-2  \\
Layer Interval $g$ & 8 \\
Graph Dimension & 512 \\
Cross-Atten LR & 5e-6 \\
\bottomrule
\end{tabular}
\end{subtable}
\end{table}

For the data used in SFT, we show an example below in Fig~\ref{fig:SFT_example}. In this setup, the Reasoning Response explicitly states the dependency structure required for reasoning.

\begin{figure*}[h]
\centering
\begin{tcolorbox}[
  enhanced,
  width=\textwidth,
  boxrule=0.8pt,
  arc=2mm,
  top=8pt,
  bottom=6pt,
  left=8pt,
  right=8pt
]

\textbf{SFT Data Example.}
\tcbline
\vspace{0.4em}
\textbf{Instruction:}
You will be given a logical reasoning question with context information. Analyze the given information carefully and determine the correct answer from the provided options.

\textbf{\textcolor{blue!70!black}{Context:}}\\
Brennan is creative. \\
Brennan does not have obstacles. \\
Brennan meets adversity. \\
If Brennan is creative and faces challenges, then he produces inspirational work. \\
Anyone who has obstacles or meets adversity faces challenges. \\

\textbf{Question:}\\
Based on the above information, is the following statement true, false, or uncertain?
\textit{Brennan does not produce inspirational work.}

\noindent\textbf{Answer:}
\textbf{B}

\vspace{0.5em}
\noindent\textbf{Reasoning Response (SFT Label)}
Brennan does not have obstacles (fact 1) but he does meet adversity (fact 2). According to the rule, anyone who has obstacles or meets adversity faces challenges. Therefore, since Brennan meets adversity, he faces challenges.\texttt{\textbackslash n\textbackslash n}Additionally, it is stated that Brennan is creative (fact 1) and that he faces challenges (fact 2). The rule indicates that if Brennan is creative and faces challenges, then he produces inspirational work. However, the conclusion drawn is that Brennan does not produce inspirational work.\texttt{\textbackslash n\textbackslash n}This leads to a contradiction, as it is false to assert that Brennan does not produce inspirational work given the conditions established. \texttt{\textbackslash n\textbackslash n} Therefore, the correct option is: B.
\end{tcolorbox}
\caption{An example of SFT training data, where the reasoning response is used as the training label to explicitly teach the LLM the dependency structure required for reasoning.}
\label{fig:SFT_example}
\end{figure*}

\subsection{Rules of Replacement}

We list below all rules of replacement used to generate logically equivalent variants.

\begin{table}[htbp]

\caption{\small {Rules of replacement used in our transformations and their corresponding tautologies.}}
\vspace{10pt}
\label{Tab:rule_demo}

\centering
\small
\setlength{\tabcolsep}{5pt}
\renewcommand{\arraystretch}{1.15}
\begin{tabular}{p{0.26\textwidth} p{0.24\textwidth}}
\toprule
\textbf{Rule of Replacement} & \textbf{Tautology} \\
\midrule
% \textsc{Commutative 1} & $(p \land q) \Leftrightarrow (q \land p)$ \\
% \textsc{Commutative 2} & $(p \lor q) \Leftrightarrow (q \lor p)$ \\
\textsc{Material implication} & $(p \rightarrow q) \Leftrightarrow (\neg p \lor q)$ \\
\textsc{Double negation intro} & $p \Leftrightarrow \neg\neg p$ \\
\textsc{De Morgan's law 1} & $\neg(p \land q) \Leftrightarrow (\neg p) \lor (\neg q)$ \\
\textsc{De Morgan's law 2} & $\neg(p \lor q) \Leftrightarrow (\neg p) \land (\neg q)$ \\
\bottomrule
\end{tabular}

\end{table}

\subsection{Robustness of LoGIST under Practical Setting}

\label{appendix:robust_pratical}

In practice, a formal representation is not always available and may need to be parsed from the natural-language input, potentially resulting in imperfect ROBDDs due to parsing errors such as mismatched predicate names, missing or extra literals, or incorrect connectives. We therefore directly study this realistic setting to evaluate LoGIST's effectiveness under more challenging conditions.

We conduct an additional experiment in which the formal representation at evaluation time is instead extracted automatically from the natural-language text, following NL2Logic~\cite{putra-etal-2026-nl2logic}. Notably, the extraction is performed by the vanilla Llama-3.1-8B-Ins model, the same base model used for training, rather than any specialized logic model. The parsed FOL is genuinely noisy: on the Mixed data, only $15.3\%$ of the parsed representations exactly match the gold FOL (mostly because predicate names differ from those in the gold annotations), while $73.6\%$ are structurally correct. The model is trained following the pipeline described in the paper and evaluated with ROBDDs built from this noisy parsed FOL. The results are shown in Tab.~\ref{tab}.

\begin{table}[t]
\centering
\caption{Performance comparison across different settings.}
\label{tab:logist_rubustness_results}
\begin{tabular}{lccccc}
\toprule
\textbf{Method} & \textbf{Base} & \textbf{M.I.} & \textbf{D.N.} & \textbf{D.M.} & \textbf{Mixed} \\
\midrule
GRPO + LoGIST (Gold FOL)        & 0.730 & 0.654 & 0.661 & 0.647 & 0.668 \\
GRPO + LoGIST (Constructed FOL) & 0.722 & 0.649 & 0.650 & 0.645 & 0.656 \\
Vanilla GRPO                    & 0.702 & 0.608 & 0.612 & 0.600 & 0.627 \\
\bottomrule
\end{tabular}
\end{table}

Our results show that even under this realistic noise level, LoGIST still yields stable improvements: when the ROBDD is imperfect, the resulting performance change is negligible and the model stays close to its gold-FOL input performance. This indicates that LoGIST remains effective with realistic, automatically parsed representations and does not rely on noise-free annotations.

\subsection{Study: Contrastive Training Enables Generalizable Logical Structure Learning }

\begin{wraptable}{r}{0.56\textwidth}
\vspace{-0.4cm}
\caption{\small{Performance comparison with and without contrastive learning for \texttt{Qwen2.5-3B-Ins} trained with GRPO.}}
\label{tab:ablation}
\vspace{-0.4em}
\centering
\footnotesize
\setlength{\tabcolsep}{3.0pt}
\renewcommand{\arraystretch}{1.15}
\resizebox{\linewidth}{!}{
\begin{tabular}{lccccc}
\toprule
\textbf{Methods} & \textbf{Base} & \textbf{Mat.Impl.} & \textbf{Double Neg.} & \textbf{De Morgan} & \textbf{Mixed}  \\
\midrule

\textbf{w/} Contrastive        & 0.655 & 0.604  & 0.607  & 0.598  & 0.620 \\
\textbf{w/o} Contrastive      & 0.651 & 0.571 & 0.584 & 0.578 & 0.595 \\
GRPO Baseline      & 0.645 & 0.570 & 0.577 & 0.572 & 0.590 \\
\bottomrule
\end{tabular}
}
\vspace{-0.8em}
\end{wraptable}

We finally examine how contrastive training enables the logic encoder to capture structural information that supports generalizable reasoning. To isolate its effect, we compare three variants that differ only in encoder initialization: \textbf{(i)} LoGIST with the contrastively trained encoder from Sec.~\ref{subsec:encoder_train}, \textbf{(ii)} LoGIST with a randomly initialized encoder (no contrastive training), and \textbf{(iii)} the GRPO baseline. As shown in Tab.~\ref{tab:ablation}, LoGIST without contrastive training yields only marginal gains over the GRPO baseline, whereas the contrastively trained encoder produces consistent and substantial improvements across all settings. This comparison indicates that the structural information needed for generalizable reasoning is successfully acquired during contrastive training, and the resulting logic embeddings encode reusable structural features that the LLM can leverage to generalize across logically equivalent inputs.

\subsection{Limitation and Potential Impact}

\label{appendix:limitation}

In this work, we unveil a key limitation of current LLMs in capturing logical structure when performing reasoning tasks, and present a framework showing that explicitly encoding logical structure can improve model reasoning generalization. This finding has potential impact for applications that require faithful symbolic reasoning, such as formal verification, theorem proving assistance, program analysis, and high-stakes decision support where consistency under equivalent logical formulations is important.

Despite these promising results, our study has limitations. First, we focus on formal reasoning in this work; extending LoGIST to domains such as code reasoning, where structural factors go beyond Boolean equivalence \textit{e.g.}, data-flow dependencies and variable scoping, remains future work. Second, our current implementation prioritizes feasibility and empirical validation; integrating LoGIST into mainstream inference engines such as vLLM and SGLang for higher-throughput deployment remains future engineering work.

\clearpage

\subsection{Prompt Template}

This section presents the basic prompt used in the paper. 

\begin{figure*}[h]
\centering
\begin{tcolorbox}[
  enhanced,
  width=\textwidth,
  colback=orange!10!white,
  colframe=brown!70!black,
  coltitle=white,
  colbacktitle=brown!70!black,
  title={\textbf{Prompt Template for Logical Reasoning}},
  fonttitle=\bfseries\large,
  boxrule=1pt,
  arc=2.5mm,
  outer arc=2.5mm,
  top=10pt,
  bottom=8pt,
  left=10pt,
  right=10pt
]
\small

\textbf{Instruction:}
You will be given a logical reasoning question with context information. Analyze the given information carefully and determine the correct answer from the provided options.

\vspace{0.5em}
\textbf{\textcolor{blue!70!black}{Context:}}\\
\textcolor{blue!70!black}{\texttt{<CONTEXT>}}

\vspace{0.8em}
\textbf{Question:}\\
Based on the above information, is the following statement true, false, or uncertain? \\

\textcolor{red!75!black}{\texttt{<STATEMENT>}}

\vspace{0.8em}
\textbf{Options:}\\
A) True\\
B) False\\
C) Uncertain

\vspace{0.8em}

Let's think step by step:

\end{tcolorbox}
\end{figure*}

\begin{figure*}[h]
\centering
\begin{tcolorbox}[
  enhanced,
  width=\textwidth,
  colback=orange!10!white,
  colframe=brown!70!black,
  coltitle=white,
  colbacktitle=brown!70!black,
  title={\textbf{Linearized ROBDD Prompt Template for Logical Reasoning}},
  fonttitle=\bfseries\large,
  boxrule=1pt,
  arc=2.5mm,
  outer arc=2.5mm,
  top=10pt,
  bottom=8pt,
  left=10pt,
  right=10pt
]
\small

\textbf{Instruction:}\\
You will be given contextual information, a linearized reasoning graph, and a query statement. Carefully analyze the provided information and determine whether the statement is \textbf{true}, \textbf{false}, or \textbf{uncertain}.

\vspace{0.8em}

\textbf{\textcolor{blue!70!black}{Context:}}\\
\textcolor{blue!70!black}{\texttt{\detokenize{<CONTEXT>}}}

\vspace{0.8em}

\textbf{\textcolor{blue!70!black}{Reasoning Graph (Linearized ROBDD):}}\\
\texttt{\detokenize{Root: N0}}\\
\texttt{\detokenize{N0: if contribute_to_film_preservation(Aurora) then N2 else FALSE}}\\
\texttt{\detokenize{N2: if appreciate_film_history(Aurora) then N3 else FALSE}}\\
\texttt{\detokenize{N3: if collect_videodisks(Aurora) then FALSE else N4}}\\
\texttt{\detokenize{N4: if study_cinema(Aurora) then TRUE else FALSE}}

\vspace{0.8em}

\textbf{Question:}\\
Based on the context and reasoning graph above, is the following statement \textbf{true}, \textbf{false}, or \textbf{uncertain}?

\vspace{0.4em}

\textcolor{red!75!black}{\texttt{\detokenize{<STATEMENT>}}}

\vspace{0.8em}
\textbf{Options:}\\
A) True\\
B) False\\
C) Uncertain

\vspace{0.8em}

Let's think step by step:

\end{tcolorbox}

\caption{Linearized ROBDD prompt template for logical reasoning.}
\label{fig:linearized_robdd}
\end{figure*}

We also provide the prompt used to guide LLMs toward recognizing equivalent logical structures, as described in Sec.~\ref{para:icl}.

\begin{figure*}[htbp]
\centering
\begin{tcolorbox}[
  enhanced,
  width=\textwidth,
  colback=orange!10!white,
  colframe=brown!70!black,
  coltitle=white,
  colbacktitle=brown!70!black,
  title={\textbf{Prompt Template for Learning Logically Equivalent Structures}},
  fonttitle=\bfseries\large,
  boxrule=1pt,
  arc=2.5mm,
  outer arc=2.5mm,
  top=8pt,
  bottom=6pt,
  left=8pt,
  right=8pt
]
\small

\textbf{Instruction.}
You will be given a logical reasoning question with contextual information.
Analyze the information carefully and determine the correct answer from the provided options.

\vspace{0.6em}
\textbf{Hints.}

\textbf{Commutative Laws}
\begin{itemize}[leftmargin=1.4em, itemsep=0.2em, topsep=0.3em, parsep=0pt, partopsep=0pt]
    \item Use this law when the same terms appear in a different order within a conjunction or disjunction.
    \item \(A \land B \Leftrightarrow B \land A\).
    \item \(A \lor B \Leftrightarrow B \lor A\).
\end{itemize}

\textbf{Material Implication Law}
\begin{itemize}[leftmargin=1.4em, itemsep=0.2em, topsep=0.3em, parsep=0pt, partopsep=0pt]
    \item Use this law when rewriting an if--then statement as an equivalent disjunction.
    \item \(A \rightarrow B \Leftrightarrow \neg A \lor B\).
\end{itemize}

\textbf{Double Negation Elimination}
\begin{itemize}[leftmargin=1.4em, itemsep=0.2em, topsep=0.3em, parsep=0pt, partopsep=0pt]
    \item Use this law when a proposition contains two consecutive negations.
    \item \(\neg\neg A \Leftrightarrow A\).
\end{itemize}

\textbf{De Morgan's Laws}
\begin{itemize}[leftmargin=1.4em, itemsep=0.2em, topsep=0.3em, parsep=0pt, partopsep=0pt]
    \item Use this law when a negation applies to an entire conjunction or disjunction and needs to be pushed inward.
    \item \(\neg(A \land B) \Leftrightarrow \neg A \lor \neg B\).
    \item \(\neg(A \lor B) \Leftrightarrow \neg A \land \neg B\).
\end{itemize}

\textbf{Key Insight.}
Apply these equivalence laws only when they help simplify the context and support the derivation of the correct answer.

\vspace{0.5em}
\textbf{\textcolor{blue!70!black}{Context.}}\\
\textcolor{blue!70!black}{\texttt{<CONTEXT>}}

\vspace{0.6em}
\textbf{Question.}\\
\texttt{<QUESTION>}

\vspace{0.6em}
\textbf{Options:}\\
A) True\\
B) False\\
C) Uncertain

\vspace{0.6em}
\textbf{Answer.} Let's think step by step.

\end{tcolorbox}
\caption{Prompt template used for prompting equivalent logical learning.}
\label{fig:icl_example}
\end{figure*}

\clearpage

\subsection{Data Example of Equivalent Logic}
\label{appendix:data_example}

In this part, we present an example of our Equivalent Logic transformation. We start by constructing the initial data using implicational rule templates, and then apply a rule of replacement to the rule set in the premise to obtain a logically equivalent variant. In this process, the transformed version shares the same fact set as the initial one, while the transformed rule set is presented to the LLM.

\begin{figure*}[htbp]
\centering
\begin{tcolorbox}[
  enhanced,
  width=\textwidth,
  boxrule=0.8pt,
  arc=2mm,
  top=8pt,
  bottom=6pt,
  left=8pt,
  right=8pt
]
\small

\textbf{Initial Facts and Rules}
\begin{itemize}[leftmargin=1.2em, itemsep=2pt, topsep=3pt]
    \item[-] Aurora does not collect videodisks. {\color{blue!70!black}$\neg \mathrm{collect\_videodisks}(\mathrm{Aurora})$}
    
    \item[-] Aurora studies cinema. {\color{blue!70!black}$\mathrm{study\_cinema}(\mathrm{Aurora})$}
\end{itemize}

\begin{itemize}[leftmargin=1.2em, itemsep=2pt, topsep=3pt]
    \item[-] If Aurora collects videodisks or appreciates film history, then she contributes to film preservation. {\color{blue!70!black}$(\mathrm{collect\_videodisks}(\mathrm{Aurora}) \vee \mathrm{appreciate\_film\_history}(\mathrm{Aurora})) \rightarrow \mathrm{contribute\_to\_film\_preservation}(\mathrm{Aurora})$}
    
    \item[-] Anyone who watches classics or studies cinema can appreciate film history. {\color{blue!70!black}$\forall x \big((\mathrm{watch\_classics}(x) \vee \mathrm{study\_cinema}(x)) \rightarrow \mathrm{appreciate\_film\_history}(x)\big)$}
\end{itemize}

\tcbline

\textbf{Material Implication}
\begin{itemize}[leftmargin=1.2em, itemsep=2pt, topsep=3pt]
    \item[-] It is not the case that Aurora collects videodisks or appreciates film history, or she contributes to film preservation. {\color{blue!70!black}$\neg(\mathrm{collect\_videodisks}(\mathrm{Aurora}) \vee \mathrm{appreciate\_film\_history}(\mathrm{Aurora})) \vee \mathrm{contribute\_to\_film\_preservation}(\mathrm{Aurora})$}
    
    \item[-] For all people, it is not the case that they watch classics or study cinema, or they appreciate film history.{\color{blue!70!black}$\forall x \big(\neg(\mathrm{watch\_classics}(x) \vee \mathrm{study\_cinema}(x)) \vee \mathrm{appreciate\_film\_history}(x)\big)$}
\end{itemize}

\tcbline

\textbf{De Morgan's Law}
\begin{itemize}[leftmargin=1.2em, itemsep=2pt, topsep=3pt]
    \item[-] If Aurora does not both fail to collect videodisks and fail to appreciate film history, then she contributes to film preservation. {\color{blue!70!black}$\neg\big(\neg \mathrm{collect\_videodisks}(\mathrm{Aurora}) \wedge \neg \mathrm{appreciate\_film\_history}(\mathrm{Aurora})\big) \rightarrow \mathrm{contribute\_to\_film\_preservation}(\mathrm{Aurora})$}
    
    \item[-] For all people, if they do not both fail to watch classics and fail to study cinema, then they can appreciate film history. {\color{blue!70!black}$\forall x \big(\neg(\neg \mathrm{watch\_classics}(x) \wedge \neg \mathrm{study\_cinema}(x)) \rightarrow \mathrm{appreciate\_film\_history}(x)\big)$}
\end{itemize}

\tcbline

\textbf{Double Negation Intro}
\begin{itemize}[leftmargin=1.2em, itemsep=2pt, topsep=3pt]
    \item[-] If it is not the case that Aurora fails to either collect videodisks or appreciate film history, then she contributes to film preservation. {\color{blue!70!black}$\neg\neg\big(\mathrm{collect\_videodisks}(\mathrm{Aurora}) \vee \mathrm{appreciate\_film\_history}(\mathrm{Aurora})\big) \rightarrow \mathrm{contribute\_to\_film\_preservation}(\mathrm{Aurora})$}
    
    \item[-] For all people, if it is not the case that they fail to either watch classics or study cinema, then they can appreciate film history. {\color{blue!70!black}$\forall x \big(\neg\neg(\mathrm{watch\_classics}(x) \vee \mathrm{study\_cinema}(x)) \rightarrow \mathrm{appreciate\_film\_history}(x)\big)$}
\end{itemize}

\tcbline

\textbf{Mixed Transformation}
\begin{itemize}[leftmargin=1.2em, itemsep=2pt, topsep=3pt]
    \item[-] If Aurora collects videodisks or appreciates film history, then she contributes to film preservation. {\color{blue!70!black}$(\mathrm{collect\_videodisks}(\mathrm{Aurora}) \vee \mathrm{appreciate\_film\_history}(\mathrm{Aurora})) \rightarrow \mathrm{contribute\_to\_film\_preservation}(\mathrm{Aurora})$}
    
    \item[-] For all people, if it is not the case that they fail to either watch classics or study cinema, then they can appreciate film history. {\color{blue!70!black}$\forall x \big(\neg\neg(\mathrm{watch\_classics}(x) \vee \mathrm{study\_cinema}(x)) \rightarrow \mathrm{appreciate\_film\_history}(x)\big)$}
\end{itemize}

\end{tcolorbox}
\caption{An example of logically equivalent transformations applied to a fact set and its associated rule sets. The fact set is kept fixed, while the rule set is rewritten under different equivalence transformations. The De Morgan case requires an implicit Double Negation Introduction step before De Morgan's Law can be applied, since De Morgan's Law strictly rewrites a negated conjunction or disjunction.}
\label{fig:logic_equiv_example}
\end{figure*}

\clearpage

% \clearpage
% \input{checklist.tex}
\end{document}